\documentclass[5p]{elsarticle}

\usepackage{amssymb}
\usepackage[figuresright]{rotating}
\usepackage{multicol}
\usepackage{graphicx}
\usepackage{amssymb}
\usepackage{algorithm,algorithmic}
\usepackage{mathtools}
\usepackage{mathptmx}
\usepackage{amsmath}
\usepackage{amssymb}
\usepackage{amsthm}
\usepackage{cancel}
\usepackage{blkarray}
\usepackage{listings}  
\usepackage{multirow}
\usepackage{lipsum}
\usepackage{subcaption}
\usepackage{caption}
\usepackage{url}
\usepackage{xcolor}
\usepackage{svg}
\usepackage{enumitem}
\usepackage{etoolbox}
\usepackage{wrapfig}
\begin{document}

\begin{frontmatter}



% \dochead{}
%% e.g. \dochead{17th International Conference on Dynamical Processes in Excited States of Solids}

\title{CTG-KrEW: Generating Synthetic Structured Contextually Correlated Content by Conditional Tabular GAN with K-Means Clustering and  Efficient Word Embedding}

\author{Riya Samanta\fnref{label2}}
\ead{riya.samanta@iitkgp.ac.in}
\author{Bidyut Saha\fnref{label2}}
\ead{bidyutsaha@kgpian.iitkgp.ac.in}
\author{Soumya K. Ghosh\fnref{label2}}
\ead{skg@iitkgp.ac.in}
\author{Sajal K. Das\fnref{label3}}
\ead{sdas@mst.edu}

\address[label2]{Indian Institute of Technology Kharagpur, India}
\address[label3]{Missouri University of Science and Technology, USA}

% \cortext[cor1]{corresponding author}
\fntext[label2]{Equal contribution}

\begin{abstract}
Conditional Tabular Generative Adversarial Networks (CTGAN) and their various derivatives are attractive for their ability to efficiently and flexibly create synthetic tabular data, showcasing strong performance and adaptability. However, there are certain critical limitations to such models. The first is their inability to preserve the semantic integrity of contextually correlated words or phrases. For instance, `skillset' in freelancer profiles is one such attribute where individual skills are semantically interconnected and indicative of specific domain interests or qualifications. The second challenge of traditional approaches is that, when applied to generate contextually correlated tabular content, besides generating semantically shallow content, they consume huge memory resources and CPU time during the training stage. To address these problems, we introduce a novel framework, \textit{CTG-KrEW} (Conditional Tabular GAN with K-Means Clustering and Word Embedding), which is adept at generating realistic synthetic tabular data where attributes are collections of semantically and contextually coherent words. \textit{CTG-KrEW} is trained and evaluated using a dataset from Upwork, a real-world freelancing platform. Comprehensive experiments were conducted to analyze the variability, contextual similarity, frequency distribution, and associativity of the generated data, along with testing the framework's system feasibility. \textit{CTG-KrEW} also takes around 99\% less CPU time and 33\% less memory footprints than the conventional approach. Furthermore, we developed KrEW, a web application to facilitate the generation of realistic data containing skill-related information. This application, available at \url{https://riyasamanta.github.io/krew.html}, is freely accessible to both the general public and the research community.

\end{abstract}


\end{frontmatter}

% \linenumbers

\section{Introduction}
\label{intro}

In today's digital landscape, data acts as the cornerstone of innovation, akin to oil in its transformative power across industries. Yet, accessibility challenges and privacy concerns impede the realization of data-driven initiatives, particularly for burgeoning academic and nascent organizations.

A prevalent strategy for addressing data scarcity is the adoption of fully synthetic data \cite{dankar2021fake}. Synthetic data refers to fabricated datasets that mirror the structure and statistical characteristics of their original counterparts. However, while synthetic content is valuable, generating it in unstructured form may not fully suffice for data-centric applications. Structured datasets, particularly heterogeneous tabular data, emerge as pivotal assets \cite{9998482}. Widely utilized and indispensable in numerous critical and computationally demanding applications, tabular datasets excel in various aspects. They not only facilitate interpretability and support extensive feature engineering but also enable benchmarking, reproducibility, scalability, and seamless integration with existing systems.

Hence, the generation of synthetic tabular data is a significant undertaking that has fascinated scholars for a long time. Previous literature has approached the issue of treating individual columns of a table differently by creating a joint multivariate probability distribution and subsequently sampling from the observed distribution. This has been accomplished through the utilisation of Bayesian networks as well as classification and regression trees \cite{Bourou2021}. Recently, there has been a growing interest in exploring the potential of Generative Adversarial Networks (GANs) \cite{NIPS2014_5ca3e9b1} models for generating synthetic tabular data. The GAN framework comprises two primary components, namely a generator network and a discriminator network. The generator network is trained to produce synthetic data samples that exhibit a similarity to the actual data distribution. On the other hand, the discriminator network endeavours to differentiate between real and generated samples. Concurrently, the generator network endeavours to produce samples that deceive the discriminator network, while the discriminator network strives to precisely differentiate between real and synthesised samples. GANs have been extensively employed in diverse domains, including but not limited to image synthesis, style transfer, and text generation.

Despite the impressive achievements of GANs in producing unstructured data like images and text, they face notable obstacles in generating heterogeneous tabular data. The conventional GANs that predominantly function on uninterrupted noise vectors encounter difficulties in grasping the structured characteristics of tabular data and the complex relationships among its diverse attributes. Moreover, it is worth noting that the loss functions frequently employed in GANs may not be inherently compatible with the structured characteristics of tabular data, thus constraining their efficacy in this field. The outcomes of a recent survey carried out by Kaggle reveal the ubiquity of tabular data in both academic and business settings \cite{kaggle2017}. The survey also underscores the intricacies that arise from the varied types of data that are present in tables, such as numerical, categorical, time, text, and cross-table references. In addition, the variables' distributions may manifest diverse shapes, including but not limited to multimodal, long-tail, and other configurations, thereby compounding the difficulties associated with producing tabular data \cite{Xu2018}.

TGAN (Tabular GAN) \cite{Xu2018} and TableGAN \cite{park2018data} are two prominent variations of GANs that are widely utilised for generating tabular data. However, each of these models has its limitations. TGAN encounters difficulties related to mode collapse, an event that arises when the generator network cannot capture the complete data distribution and instead generates a restricted range of samples \cite{Rai2023}. Furthermore, in \cite{NEURIPS2019_254ed7d2}, the authors have observed that the transformation of discrete variables with more than four categories into continuous values using TGAN may not produce favourable outcomes. Conversely, TableGAN postulates that the input data adhere to a continuous distribution and concentrate predominantly on producing continuous features.  It may not be as effective in capturing and generating categorical features. Therefore, to address these challenges, a more definitive model was deemed necessary, leading to the development of the Conditional Tabular Generative Adversarial Network (CTGAN) \cite{NEURIPS2019_254ed7d2}. CTGAN extends the Conditional GAN \cite{mirza2014conditional} framework by integrating conditional generation. This allows for the generation of samples that are conditioned on a particular attribute and includes an inherent mechanism for managing categorical data. The model employs an embedding layer to encode categorical variables, enabling the learning of continuous representations of categorical features \cite{fang2022,Zhang2021,Bourou2021,8939081, park2018data,dhami2021beyond}.

The current CTGAN algorithms are constrained in their treatment of variable types, primarily addressing continuous and categorical variables. This oversight neglects a significant category of contextually correlated word sequences essential for capturing the structured organization and semantic cohesion found in descriptions of particular attributes, characterizations, skill sets, or qualities. Essentially, current iterations of CTGAN are unable to retain the semantic significance and correlation of words within such sequences within a tabular context. However, in various real-world dataset scenarios, such as customer reviews and ratings, social media post datasets, product descriptions and sales datasets, worker profile descriptions, job advertisements, and more, there is a crucial need to intentionally incorporate sequences of correlated words. This is essential for conveying a nuanced understanding of subjects within a tabular format. For example, the `skillset' column in worker profiles serves as one such attribute where individual skills are semantically interconnected, offering insights into specific domain interests or qualifications. 

Regrettably, in the case of research in the crowdsourcing category itself, there is hardly any availability of skill-oriented freelancer (or worker) and task profile description datasets that can be utilized in the testing and experimentation of skill-oriented task allocation problems. To our knowledge, MeetUp \cite{meetup,sethi2023scalable,Song2020,Cheng2016, samanta2024empowering} and UpWork \cite{upwork,samanta2021swill,samanta2022volunteer,samanta2024sustainable} are the only recognized datasets, both in tabular (CSV) format, extensively used for such procedures. These datasets are not accessible via any open source repository. The researchers have either scraped the data from the websites of the stakeholders or shared it on demand. However, a repository of the UpWork data has recently been made available on Kaggle \cite{crawlfeed}. This dataset was also collected by crawling the internet on January 11th, 2022, and contains only a sample of freelance jobs posted in January 2022, without any records of the workers' or freelancers' profiles. On the other hand, the UpWork data mentioned in \cite{upwork,samanta2021swill,samanta2022volunteer} includes separate CSV files for tasks and workers. One positive aspect of Kaggle's UpWork dataset \cite{crawlfeed} is that, in addition to being relatively recent, it contains 298 unique task profiles, whereas the former dataset only had 97. Some of the co-authors of this paper have access to the UpWork dataset mentioned in\cite{samanta2021swill,samanta2022volunteer}. 
% \vspace{-0.2in}

\begin{figure*}[!ht]
  \centering
  \includegraphics[width=0.9\textwidth, height=50mm]{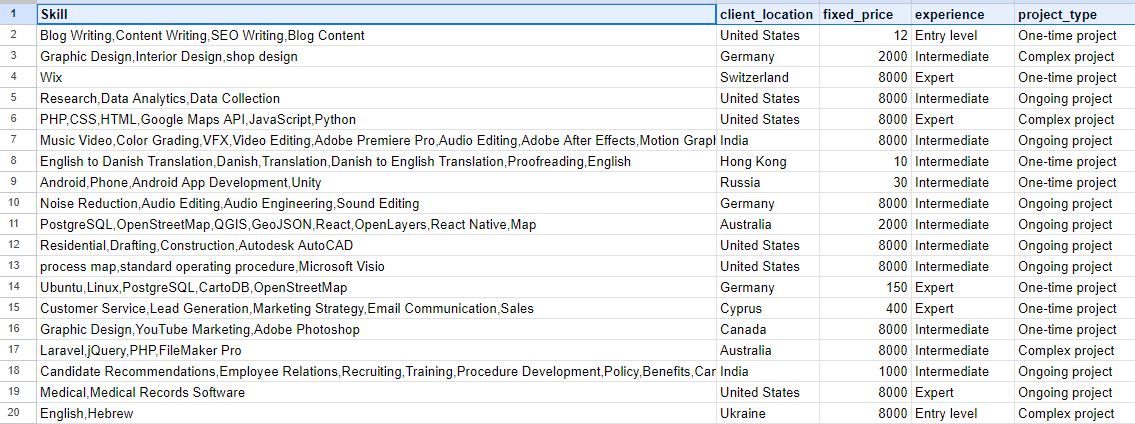}
  \vspace{-0.1in}
 \caption{Snapshots of \emph{task-data} showing only 20 rows, outsourced from Kaggle \cite{crawlfeed} }
  \label{kaggle-screen}
    \includegraphics[width=0.9\textwidth,  height=50mm]{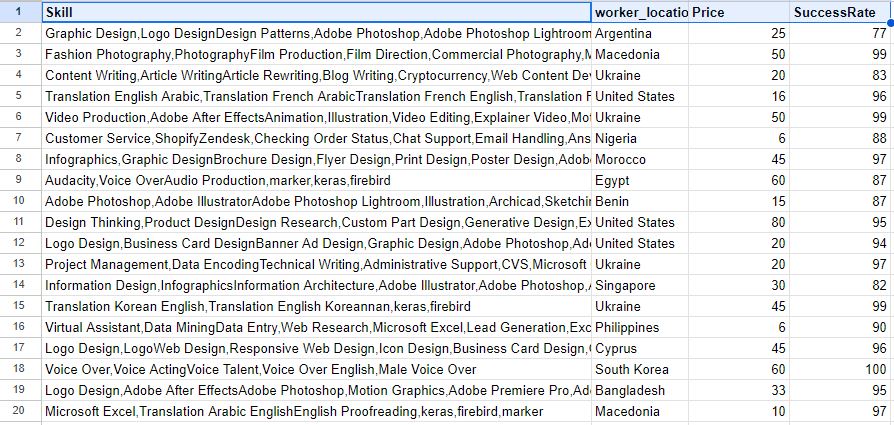}
     \caption{Snapshots of \emph{worker-data} showing only 20 rows, collected from papers \cite{samanta2021swill,samanta2022volunteer}}
      \vspace{-0.1in}
    \label{upwrk-screen}
    \vspace{-0.1in}
\end{figure*}

This study proposes a framework called CTG-KrEW (Conditional Tabular GAN with K-Means Clustering and Word Embedding) to produce extensive datasets from a limited, small-scale dataset. The \textit{CTG-KrEW}'s generated content is in a tabular context and supports continuous, categorical, and sequenced words with the semantic integrity of their contextual context. Further, the \textit{CTG-KrEW}  is designed to be system-feasible, consuming far less CPU time and memory resources than the conventional approach. \textit{CTG-KrEW} employs the word2vec method \cite{church2017word2vec} to transform individual words (e.g. skills) into vector representations, followed by K-Means clustering \cite{hartigan1979algorithm} to group them. In this study, we used task data from Kaggle's repository \cite{crawlfeed} and worker data from the aforementioned research work of \cite{samanta2021swill,samanta2022volunteer}, and subsequently referred to them as \emph{task-data} and \emph{worker-data} throughout the article. The respective snapshots of the data sets are represented in Figures \ref{kaggle-screen} and \ref{upwrk-screen}.
Instead of employing a preexisting word2vec model, we opted to train our word2vec model directly from the \emph{task-data} and \emph{worker-data} to maintain the associations among the skills that pertain to a specific skill set. We also developed a web-based application named \textbf{\textit{Krew}} that allows users to generate worker-profile and job-description-related tabular content with skill information of any magnitude. Krew is also freely accessible to the public. 

While CTG-KrEW effectively addresses the challenges of generating realistic synthetic datasets for skill-oriented profiles, its utility extends far beyond this specific application. CTG-KrEW is designed to generate generic synthetic tabular data where contextually correlated words are crucial, making it a versatile tool for any domain that requires maintaining the semantic integrity and contextual coherence of tabular data attributes. In this study, we demonstrate its capabilities using skill-oriented datasets as an example, but the framework is broadly applicable to a wide range of scenarios where realistic, contextually aware synthetic data is needed.

% To support our choice, in Section \ref{app}, we will discuss the agenda and working principle of all three variants. Subsequently, in Section \ref{eval}, a comparative analysis will be conducted among the variants, leading to the selection of \textit{\textit{CTG-KrEW}} as the ultimate model for the tool.

The rest of the paper is structured in the following manner: Section \ref{rw} provides a concise overview of the existing literature related to the generation of synthetic tabular data using GAN models. Section \ref{SOD} provides an overview of the dataset description and the challenges associated with generating synthetic tabular data of the same distribution. Section \ref{prem} presents the preliminaries required for this study. Section \ref{app} delineates the CTG-KrEW framework. In Section \ref{eval}, data from experiments are presented to assess the framework's efficacy. A description of the implementation details of the web-based application is given in Section \ref{tool}. Next is a separate discussion ( Section \ref{dis}) about the practical use cases of CTG-KrEW. The paper is concluded in Section \ref{conc}.

\vspace{-0.1in}
\section{Related Work}
\label{rw}
\vspace{-0.1in}
This section will provide a comprehensive review of the current literature on use cases in which synthetic tabular data has been generated using GAN-based approaches. 

The authors of the paper \cite{Bourou2021}, provide an extensive analysis of models based on the generative adversarial network (GAN) to synthesize tabular intrusion detection system (IDS) data. Specifically, the study uses CTGAN, TableGAN, and CopulaGAN on the widely recognized NSL-KDD dataset. Based on the analysis conducted, it can be inferred that TableGAN demonstrates satisfactory performance when handling continuous data. However, its effectiveness is limited in scenarios where discrete values are involved. In contrast, CTGAN and CopulaGAN exhibit satisfactory performance for data sets containing both continuous and discrete variables. Despite the promising capability of GAN models, they are vulnerable to various privacy attacks that could reveal information about individuals from the training data. 
The authors of \cite{fang2022} have proposed DP-CTGAN to maintain data privacy while ensuring the accuracy of the generated data. This approach involves integrating differential privacy into a conditional tabular generative model.  The authors employed nine authentic datasets, primarily sourced from the medical field, as a means of illustrating their use-case. In \cite{Zhang2021}, the authors introduced GANBLR, a GAN model, which draws inspiration from the relationship between Naive Bayes and Logistic Regression. This model is designed to overcome the interpretation limitations of current tabular GAN-based models, while also enabling the explicit handling of feature interactions. In a recent publication \cite{Rai2023}, CTGAN is employed to produce a synthetic dataset for Attribute-based Access Control (ABAC). Additionally, the researchers have developed a software tool called ConGRASS, which facilitates the creation of extensive ABAC datasets. 

However, to the best of our knowledge, none of the previously mentioned works has considered the generation of synthetic tabular datasets with attributes containing collections of words that are both highly correlated and contextually associated (similar to the skills in a skillset). \textbf{ CTG-KrEW} stands as the pioneering effort in this domain.

\begin{table*}[!ht]
\centering
\caption{Description of UpWork dataset (used in this study)}
\vspace{-0.1in}
\label{upwork}
\renewcommand{\arraystretch}{1.2}
\resizebox{\textwidth}{!}{%
\begin{tabular}{|c|cc|cc|}
\hline
& \multicolumn{2}{c|}{\textbf{task-data}} & \multicolumn{2}{c|}{\textbf{worker-data}} \\
\hline
& \multicolumn{1}{c|}{Before-preprocessing} & After-preprocessing & \multicolumn{1}{c|}{Before-preprocessing} & After-preprocessing  \\
\hline
\textbf{\#rows}     & \multicolumn{1}{c|}{298}                  &   298                  & \multicolumn{1}{c|}{1575}                 & 1575                \\ \hline
\textbf{\#coloumns} & \multicolumn{1}{c|}{15}                   & 5                   & \multicolumn{1}{c|}{8}                    & 4               \\
\hline
\textbf{Attributes removed} & \multicolumn{2}{c|}{\begin{tabular}[c]{@{}c@{}}index, url, title, description, hrs\_per\_week,\\ project\_duration, job\_type, total\_jobs\_posted\_by\_client,\\ total\_spent\_by\_client, uniq\_id, and scraped\_at \end{tabular}} & \multicolumn{2}{c|}{\begin{tabular}[c]{@{}c@{}}title, name, money\_earned,\\ and keywords\end{tabular}} \\
\hline
\textbf{Attributes used} & \multicolumn{2}{c|}{\begin{tabular}[c]{@{}c@{}}skills, client\_location, fixed\_price(budget),\\ experience, and project\_type\end{tabular}} & \multicolumn{2}{c|}{\begin{tabular}[c]{@{}c@{}}skills, worker\_location,\\ price, and job\_success\_rate\end{tabular}} \\
\hline
\end{tabular}
\vspace{-0.1in}
}
\end{table*}

\vspace{-0.15in}
\section{Dataset with Sequentially Correlated Word Attributes}
\label{SOD} 
\vspace{-0.06in}
The data used is in two main CSV files. The \emph{worker-data} file, used in previous studies \cite{samanta2021swill,Sama2212:Volunteer}, contains worker profiles from UpWork. The \emph{task-data} file is sourced from Kaggle \cite{crawlfeed} and includes information on posted tasks on the UpWork platform. Both files feature a `skills' column that lists the necessary skillsets for tasks or workers. These skills represent sequences of contextually correlated words capturing the entities' competencies and cognitive qualities (refer to Table \ref{example}). For instance, a `full-stack developer' task might require the skillset `Java, JavaScript, and HTML'. Table \ref{upwork} outlines the dataset attributes. The \emph{task-data} includes categorical attributes like `skills', client\_location', `experience', and `project\_type', and a continuous `fixed\_price' attribute. The \emph{worker-data} has categorical variables such as `worker\_location' and `Success\_Rate', and a continuous `Price' variable.

The primary challenge arises in handling the `skills'. A simplistic approach would involve treating each unique entry in the `skills' column as an independent discrete category. For instance, considering the previous example, a `full-stack-developer' task requiring ``Java, JavaScript, and HTML" would be treated as one category, while a `Data analyst' task with skills ``Python, R" would be treated as another separate category. It is known that datasets with columns having such categorical values are converted using \textit{one-hot encoding} by CTGAN. This may cause the CTGAN model to produce a considerable number of repetitive entries in the synthetic data. Consequently, skillsets such as ``Java, JavaScript, and HTML" and ``Python, R" may consistently appear as inseparable combinations, ultimately leading to a reduction in information variability and potentially causing an information imbalance within the generated data.

% \vspace{-0.05in}
\section{Preliminaries}
\label{prem}
\vspace{-0.1in}

The primary challenge of this research is to generate content in a tabular context and support continuous, categorical, and sequenced words with semantic integrity. In this section, we will discuss the agenda and framework of CTG-KrEW. 

Table \ref{example} presents seven sample entities $E=\{e_{1},e_{2},...e_{7}\}$ (which could be either tasks or workers). Each entity $e_{i} \in E$ possesses a set of skills, or more appropriately, a \textit{skillset} and is represented by $S_{i}$.  For example, entity $e_{3}$ has a skillset $S_{3} = $ ``Java, Javascript, HTML". It is worth mentioning that the length of the skillsets may vary across different entities. Furthermore, two entities can possess either identical or disjoint skillsets. For instance, we observe that $S_{2} \equiv S_{7}$, but $S_{2} \not\equiv S_{j}, \forall j \in E \setminus \{2, 7\}$.

\begin{table*}[!t]
\centering
\caption{Example case of 7 entities (either worker or task) with their skillsets}
\renewcommand{\arraystretch}{1.2}
 \setlength{\tabcolsep}{10pt} 
\label{example}
\resizebox{\textwidth}{!}{%
\begin{tabular}{|c|c|ccc|}
\hline
\multirow{2}{*}{\textbf{Entity}} &
  \multirow{2}{*}{\textbf{Skillset}} &
  \multicolumn{3}{c|}{\textbf{Dimenesion of encoded skill matrix}} \\ \cline{3-5} 
 &
   &
  \multicolumn{1}{c|}{\textbf{Generic CTGAN}} &
  \multicolumn{1}{c|}{\textbf{CTGAN with MHE}} &
  \textbf{CTG-KrEW} \\ \hline
$e_{1}$ &
  C++,C, Java &
  \multicolumn{1}{c|}{\multirow{7}{*}{\begin{tabular}[c]{@{}c@{}}$(7 \times (p+m))$\\ \\ *p= number of original columns \\ *m: number of unique skillsets\\ m=6\end{tabular}}} &
  \multicolumn{1}{c|}{\multirow{7}{*}{\begin{tabular}[c]{@{}c@{}}$(7 \times (p+n))$\\ \\ *p= number of original columns \\ *n: number of unique skills\\ n=9\end{tabular}}} &
  \multirow{7}{*}{\begin{tabular}[c]{@{}c@{}}$(7 \times (p+z))$\\ \\ *p= number of original columns\\ z*: number of skill clusters\end{tabular}} \\ \cline{1-2}
$e_{2}$ &
  HTML,Javascript &
  \multicolumn{1}{c|}{} &
  \multicolumn{1}{c|}{} &
   \\ \cline{1-2}
$e_{3}$ &
  Java, Javascript, HTML &
  \multicolumn{1}{c|}{} &
  \multicolumn{1}{c|}{} &
   \\ \cline{1-2}
$e_{4}$ &
  PHP, Javascript,HTML &
  \multicolumn{1}{c|}{} &
  \multicolumn{1}{c|}{} &
   \\ \cline{1-2}
$e_{5}$ &
  Java, PHP,Node.js &
  \multicolumn{1}{c|}{} &
  \multicolumn{1}{c|}{} &
   \\ \cline{1-2}
$e_{6}$ &
  Python, R &
  \multicolumn{1}{c|}{} &
  \multicolumn{1}{c|}{} &
   \\ \cline{1-2}
$e_{7}$ &
  HTML, Javascript &
  \multicolumn{1}{c|}{} &
  \multicolumn{1}{c|}{} &
   \\ \hline
\end{tabular}%
 }
\end{table*}

\vspace{-0.2in}
\begin{figure*}[!t]
  \centering
  \includegraphics[width=0.8\textwidth]{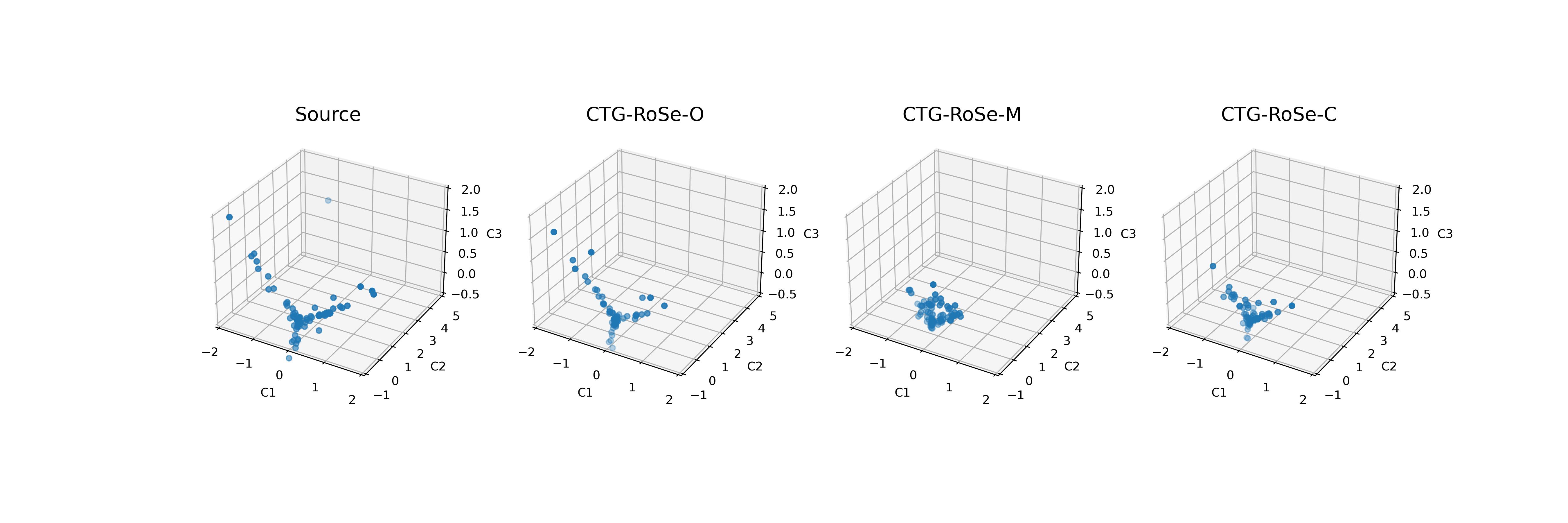}
  \vspace{-0.3in}
 \caption{Skill distribution in the \textit{task-data} of the source and synthetic datasets in 3D space with coordinates in the range $[-2,2]$, $[0,5]$, and $[-0.5,2]$ by using PCA.}
\label{fig:3d-t}
% \vspace{-0.1in}
    \includegraphics[width=0.8\textwidth]{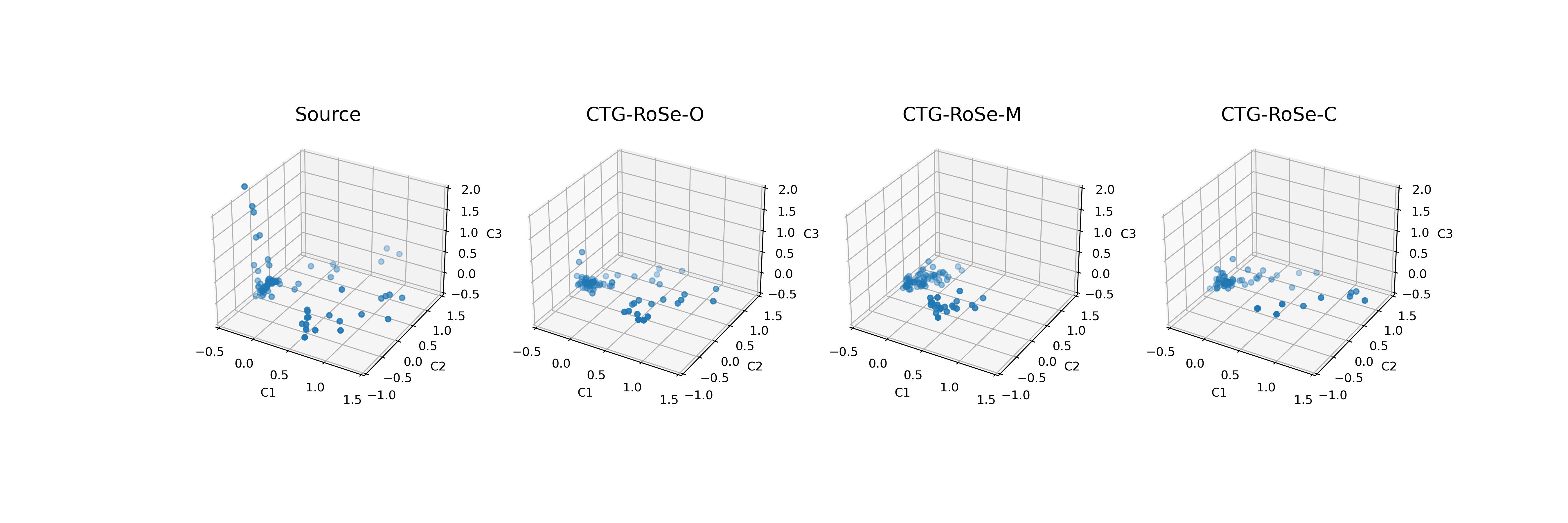}
    \vspace{-0.3in}
     \caption{Skill distribution in the \textit{worker-data} of the source and synthetic datasets in 3D space with coordinates in the range $[-0.5,1.5]$, $[-1,1.5]$, and $[-0.5,2]$ by using PCA.}
    \label{3d-w}
    % \vspace{-0.1in}
\end{figure*}

\vspace{0.1in}
\subsection{Generic CTGAN Approach with Default Encoding}
\label{ver1}
CTGAN's default approach is to encode categorical values using a \textit{one-hot encoding} technique. This involves identifying all the unique skillsets present in the dataset and creating binary columns for each distinct skillset. For instance, in Table \ref{example}, there are $m=6$ unique skillsets. Consequently, 6 binary columns, denoted as $Col_{1}, Col_{2}, \ldots, Col_{6}$, will be introduced. For a given entity $e_{i}$, the value of its corresponding $Col_{j}$ column will be set to 1 if the respective skillset $S_{j}$ represented by $Col_{j}$ is possessed by $e_{i}$. Conversely, all other binary columns associated with that entity will be assigned zero. Therefore, if the dimension of the source data frame in our cited example was $(7 \times p)$, the dimension of CTGAN's input data frame after the default transformation would be $(7 \times (p+6))$.

However, this default encoding scheme has limitations. One drawback is that it cannot generate a new set of correlated associative words (e.g. skillsets) for records that are not present in the source dataset. For instance, Table \ref{example}, depicts that there are six unique skillsets. Thus, if we aim to generate 10 records, the maximum number of unique skillsets in the generated data will not exceed six. Hence, the synthetic data will have lots of redundancies and will have less diversity. 

\subsection{Generic CTGAN Approach with Multi-hot Encoding}
\label{ver2}
An alternative encoding approach called \textit{multi-hot encoding} is proposed in place of the default one-hot encoding used in the generic CTGAN model. Multi-hot encoding (MHE) is a modification of one-hot encoding, except that the former employs a binary column for each unique word (i.e. skill) found in the dataset.  Each binary column indicates the presence or absence of a specific skill within a skillset for the respective entity. Multi-hot encoding introduces high dimensionality to the data, particularly when dealing with a large number of unique words (i.e. skills).  In the example, there are 9 unique skills: \{C++, C, Java, HTML, Javascript, PHP, Node.js, Python, R\} implying $n=9$. However, MHE introduces the additional challenge of higher training time and memory consumption.

% To address this limitation we move on to our next variant
\vspace{-0.1in}
\section{Proposed CTG-KrEW Framework}
\label{app}
\vspace{-0.1in}
\textit{CTG-KrEW} presents an innovative encoding technique that strives to incorporate greater variability in generating unique set of correlated and associative words (like skillsets) for synthetic data while simultaneously ensuring that the data dimensionality remains feasible for training CTGAN. 

The data undergoes three main preprocessing steps before training CTGAN: \textit{unique skill (or word) identification, word2Vec encoding,} and \textit{clustering}. The K-means clustering algorithm \cite{hartigan1979algorithm} is utilised by specifying a user-defined hyperparameter, represented as $K$, that establishes the intended number of cluster centres.  In contrast to prior variations discussed in subsections \ref{ver1} and \ref{ver2}, in \textit{CTG-KrEW}, new columns are introduced to represent the cluster IDs. Each $e_{i}$'s skillset $S_{i}$ is encoded so that the cluster-ID column is assigned the count of unique words (or skills) present in that specific \textit{cluster}.

The word2vec \cite{church2017word2vec} is employed for the encoding of the set of words (or skillset) before the clustering process. The word2vec methodology involves the transformation of individual skills into a vector representation. Rather than utilising any pre-trained word2vec model, we train the model directly using the provided \textit{task-data}. To construct the corpus utilised for training the 
word2vec model, a distinct keyword (represented as $tag_{j}$) is incorporated both preceding and succeeding each skill within a given skillset. Table \ref{tagged} displays the transformed corpus utilised to train the word2vec model, following the example illustrated in Table \ref{example}.

\begin{table}[ht]
\centering
% \captionsetup{font=small}
\caption{Example corpus for training word2vec model}
\label{tagged}
\renewcommand{\arraystretch}{1.5}
\begin{tabular}{c}
\hline
 tag0, C, tag0, C++, tag0, Java, tag0\\
 tag1, HTML, tag1, JavaScript, tag1 \\

tag2, Java, tag2, JavaScript, tag2, HTML, tag2 \\
 tag3, PHP, tag3, JavaScript, tag3, HTML, tag3 \\
tag4, Java, tag4, PHP, tag4, Node.js, tag4 \\
tag5, Python, tag5, R, tag5 \\
 tag6, HTML, tag6, JavaScript, tag6 \\
\hline
\end{tabular}
\end{table}

Additionally, during the training of the word2vec model, a window size of unity is chosen. This ensures that skills within the same skill set are positioned closer to each other in terms of distance. As a result, the likelihood of these skills being assigned to the same cluster after the clustering process increases. This approach aims to preserve the associations among the skills belonging to a specific skillset. The algorithm \ref{algo1} depicts the pseudo-code of \textit{CTG-KrEW} method.

Continuing with the previous example, after applying K-means clustering to the corpus presented in Table \ref{tagged}, consisting of 9 unique skills, 4 clusters are obtained: $Cul_{1}=$(Python, R), $Cul_{2}=$ (HTML, Javascript), $Cul_{3}=$ (C++, C, Java), and $Cul_{4}=$ (PHP, Node.js). The encoded clustered representation of these clusters is provided in Table \ref{encoded}.

\vspace{-0.1in}
\begin{table}[ht]
\centering
% \captionsetup{font=small}
\caption{Transformed dataset for CTGAN input}
\label{encoded}
% \vspace{-0.1in}
\renewcommand{\arraystretch}{1.2}
\setlength{\tabcolsep}{4pt}
\begin{tabular}{|c|c|c|c|}
\hline
$Cul_{1}$ & $Cul_{2}$ & $Cul_{3}$ & $Cul_{4}$ \\
\hline
0         & 0         & 3         & 0         \\
\hline
0         & 2         & 0         & 0         \\
\hline
0         & 2         & 1         & 0         \\
\hline
0         & 2         & 0         & 1         \\
\hline
0         & 0         & 1         & 2         \\
\hline
2         & 0         & 0         & 0         \\
\hline
0         & 2         & 0         & 0         \\
\hline
\end{tabular}
\vspace{-0.1in}
\end{table}

\vspace{0.1in}
The analysis of Tables \ref{example} and \ref{encoded} reveals that in the case of $e_{1}$, the three skills (C++, C, Java) are categorized under $Cul_{3}$. Consequently, in Table \ref{encoded}, the initial row about $Cul_{3}$ displays a value of three. Similarly, in terms of $e_3$, HTML and JavaScript fall into the category of $Cul_2$, while Java is classified under $Cul_3$. Hence, the entry in Table \ref{encoded} about $Cul_{2}$ is two, while that of $Cul_{3}$ is one. This cluster-encoded representation form, as presented in Table \ref{encoded}, is used as input for the training of our CTGAN model. Upon completion of the training process, the model can produce synthesized data, although in its encoded form. Therefore, to ensure that the synthesized data bears a resemblance to the source dataset a \textit{decoding} function was incorporated into the \textit{CTG-KrEW} framework.

During the encoding stage, alongside the allocation of cluster IDs to individual skills, probability values are also assigned to reflect the probability that a skill is present within a given cluster (see Steps 8 to 17 of Algorithm \ref{algo1}). The decoding process uses these probability values linked to the cluster IDs to reconstruct the `skills' column, thus mapping the cluster IDs to their respective skillsets. This process enables the reconstruction of the initial skillsets for each entity, ultimately allowing for the synthesised data to be converted from its encoded form to a format that resembles the source dataset. The data produced hold the original structure and integrity of the `skills' attribute.

A visualization analysis (refer to Figures \ref{fig:3d-t} and \ref{3d-w}) was performed to compare the source data set with the synthetically generated data produced by three variants. Our analysis centered on the values within the `skills column. The datasets possess a high-dimensional nature due to the variable number of unique skills they contain. To improve the clarity of the data, principal component analysis (PCA) was used to project the high-dimensional data onto a three-dimensional space \cite{An2021,hasan2021review}. To generate the three-dimensional projections, a sample size of 90 was used for each variant of the model, and the entities are randomly sampled from their respective data sets. Subsequently, PCA was employed on the source data to extract its fundamental structure and patterns. The same PCA transformation was then implemented on the synthesized data sets, thus aligning them with the established structure of the source dataset.

The generic CTGAN (refer subsection \ref{ver1}) and CTGAN with MHE (refer subsection \ref{ver2}) are considered the baselines for this study.

% The Figures \ref{fig:3d-t} and \ref{3d-w} depict skillsets that are derived from the `skills' column. The similarities or differences that exist between the synthesised datasets and the source data in a succinct and significant manner are highlighted in the illustrations.

\subsection{Core Architecture of CTGAN }
The implementation of CTGAN for all three variants was facilitated through the utilisation of the library offered by \textit{SDV-Synthetic Data Vault}, an open-source ecosystem of libraries designed for synthetic data generation \cite{SDV,Xu2018}.

% Presented herein is a concise discussion of the architectural design implemented in the SDV CTGAN module.

CTGAN utilises a conditional generator denoted as \(G(V, Z)\), where \(V\) represents the conditional vector obtained through the one-hot encoding of discrete columns, and \(Z\) is a vector of random noise and a training-by-sampling technique to ensure equitable representation of feasible values for discrete attributes in the training stage. \(G(V, Z)\) is designed to produce a replica of the conditional vector by introducing random noise. This is achieved by minimising the generator loss ($L_G$ ) by computing the cross-entropy between the input conditional vector and the generated vector. The discriminator ($D(S)$) assesses the data samples $S$ (either real or generated) through Discriminator Loss ($L_D$) by evaluating the distance between the learned conditional distribution of the generated samples and the conditional distributions of real data. Fully connected networks are utilised in both the generator and the discriminator to capture correlations among columns. 
The Generator \(G\) consists of two fully connected hidden layers with batch normalization and ReLU activation functions:
\begin{equation*}
 G(V, Z) = \tanh(W_2 \cdot \text{ReLU}(W_1 \cdot [V, Z] + b_1) + b_2 )
\end{equation*}
where \(W_1\) and \(W_2\) are weight matrices, \(b_1\) and \(b_2\) are bias vectors, and \([V, Z]\) concatenates the one-hot encoded conditional vector \(V\) and random noise \(Z\). Synthetic data row \(S\) is generated using various activation functions, including scalar values from the hyperbolic tangent function (\(\tanh\)), mode indicators, and discrete values from the softmax function.

% The training procedure iteratively updates the generator and discriminator models in a mutually exclusive manner.

The utilisation of adversarial training methodology yields a generator that is capable of generating synthetic data through the implementation of Gumbel softmax. In contrast, the discriminator employs leaky ReLU functions and applies dropout regularisation to every hidden layer. The implementation of the PacGAN framework with 10 samples per pac is utilised as a measure to mitigate mode collapse.

The loss functions for training are defined as follows:
\begin{align*}
L_D & = -\mathbb{E}\left[S \sim \mathbb{P}_{\text{real}}\left[D(S)\right]\right] \\    &\quad + \mathbb{E}\left[S' \sim \mathbb{P}_{\text{generated}}\left[D(S')\right]\right] \\    &\quad + \text{Gradient Penalty} \\L_G & = -\mathbb{E}\left[S' \sim \mathbb{P}_{\text{generated}}\left[D(S')\right]\right]
\end{align*}
The training of the model is carried out by utilising the Wasserstein loss function, which is augmented with a gradient penalty.  \(S\) typically represents real data samples drawn from the true data distribution \(\mathbb{P}_{\text{real}}\) and \(S'\) represents generated data samples produced by the generator \(\mathbb{P}_{\text{generated}}\). Additionally, the Adam optimizer is utilised with a learning rate of $2 \times 10^{-4}$.

\begin{figure}[!t]
  \centering
  \includegraphics[width=0.50\textwidth,height=70mm]{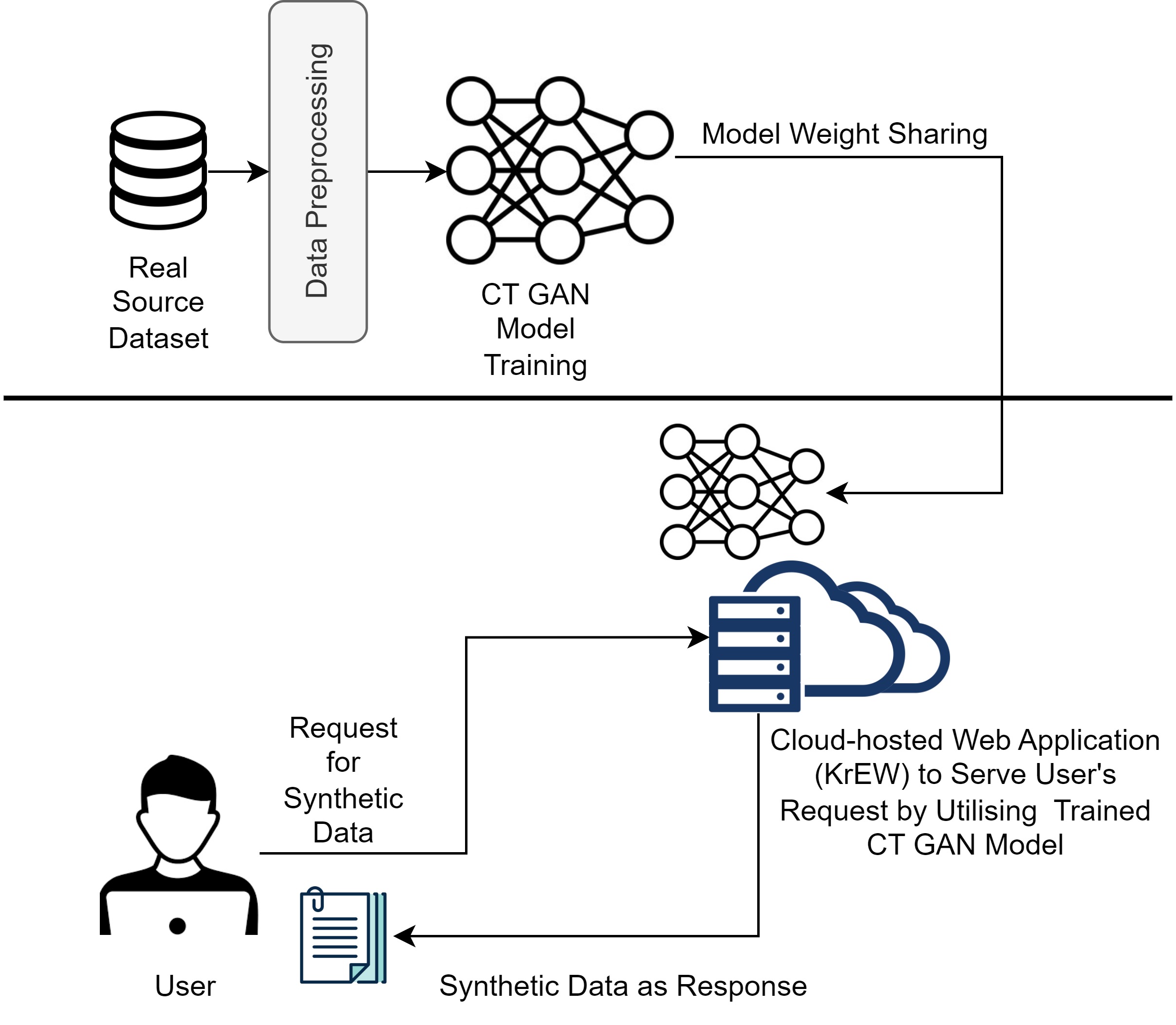}
 \caption{Workflow of \textit{CTG-KrEW}}
 \label{workflow}
\end{figure}

Figure \ref{workflow} outlines the CTG-KrEW workflow, beginning with data preprocessing, where unique skills are identified, encoded using word2vec, and clustered via K-Means. The CTGAN model then trains on this processed data to generate synthetic datasets that maintain the original data's contextual integrity. The synthetic data is then decoded back into its original format. Finally, the KrEW application, deployed on a server, enables users to generate and download these datasets at any scale.

\vspace{-0.1in}
\section{Evaluation}
\label{eval}
\vspace{-0.1in}
The proposed \textit{CTG-KrEW} and the baselines will be compared for their effectiveness using statistical and visual metrics in this study. The subsequent are the evaluation criteria with related metrics.

\vspace{0.1in}
\noindent(i) \textbf{Skillset variability:} In order to capture the variability or randomness of the \textit{skillset} values in the synthetic dataset, \textbf{\textit{Entropy}} is used. 
The concept of entropy in information theory \cite{shannon1948mathematical,mackay2003information} pertains to the mean degree of \textit{information} or \textit{uncertainty} intrinsic to the potential outcomes of a random variable.  

The distinctive skillsets are identified and their respective frequencies are recorded from both the source and synthetic datasets. As demonstrated in Table \ref{example}, the count of distinct sets of skills is $m=6$. Subsequently,  Table \ref{entropy} is generated for the skillsets. The formula for the entropy of a discrete random variable X, which is defined over a set of values ${\mathcal {X}}$ and follows a distribution function ${\displaystyle p:{\mathcal {X}}\to [0,1]}$ given by:
\begin{equation*}
{\displaystyle \mathrm {H} (X):=-\sum _{x\in {\mathcal {X}}}p(x)\log p(x)=\mathbb {E} [-\log p(X)]} 
\end{equation*}
$X$ is the normalised frequency of distinct skillsets. As the value of ${\displaystyle \mathrm {H} (X)}$ increases, the variability of skillsets also increases.

\begin{algorithm} [!t]
 \caption{CTG-KrEW} \label{algo1}
 \algsetup{linenosize=\tiny}
  \scriptsize
 \begin{algorithmic}[1]
 \renewcommand{\algorithmicrequire}{\textbf{Input:}}
 \renewcommand{\algorithmicensure}{\textbf{Output:}}
 \REQUIRE Source dataset $D$, Number of rows to generate $num\_rows$
 \ENSURE Synthesized dataset $D^{'}$ with preserved structure and integrity
 \\
% \textbf{Preprocessing:}
\STATE Initialise $mapper \gets dict()$, $sumFreq,row \gets 0$ 
\STATE  $skillsets \gets $ Extract the `skills' column from $D$
% \FOR{each $record \in skillsets$}
% \STATE Trim white space from each skill in $record$
% \ENDFOR
\STATE Construct $corpus$ by appending keyword $tag$ as shown in Table \ref{tagged}
\STATE Train a word2vec model from custom $corpus$: word2vec.$fit(corpus)$
\STATE Perform unique skill identification from $corpus$: $U \gets \text{uniqueSkills}(corpus)$\;
\STATE Find vector embedding of $U$: $encodedSkills \gets \text{word2vec}.encode(U)$
\STATE Apply K-means clustering algorithm with parameter $K$, where $K$ has been determined by \textit{elbow method}: $cluster \gets \text{KMeans}(encodedSkills, K)$\;
\FOR{$i=1$ to $length(cluster)$}
\STATE $groupedSkills \gets$ Find the list of skills assigned in $Cul_{i}$
\FOR{each $skill$ in $groupedSkills$ }
\STATE Find the count of occurrence of $skill$ in $corpus$: $freqSkill[skill] \gets \text{countOccurence}(skill,corpus)$
\STATE $sumFreq \gets sumFreq+ freqSkill[skill]$
\ENDFOR
\FOR{each $skill$ in $groupedSkills$ }
\STATE $membership[skill] \gets \frac{freqSkill[skill]}{sumFreq} $
\ENDFOR
\STATE $mapper[Cul_{i}]\gets \{groupedSkills,membership\}$
\ENDFOR
\STATE Create a table $interimTable$ with columns to represent cluster IDs \COMMENT{Refer Table \ref{encoded}}
\FOR{each $record$ $\in skillsets$}
\STATE $row \gets row+1$
\FOR{each $skill \in record$}
\FOR{$i=1$ to $length(cluster)$}
\IF{$skill$ belongs to $mapper[Cul_{i}][0]$}
\STATE $interimTable[row][Cul_{i}] \gets interimTable[row][Cul_{i}]+1$
\ENDIF
\ENDFOR
\ENDFOR
\ENDFOR
% \STATE Remove all other columns from the $encodedSkills$ data frame except those with cluster IDs: $clusteredSkillData \gets \text{clean}(encodedSkills)$ \COMMENT{Refer to Table \ref{encoded}}
\STATE CTGAN\_synthesizer$.fit(interimTable)$
\STATE CTGAN\_synthesizer$.save()$
\STATE $synthesizer \gets$ CTGAN\_synthesizer$.load()$
\STATE $D^{'} \gets$ CTGAN\_synthesizer$.sample(num\_rows)$
\STATE Create a new column $synthetic\_skill$ in $D^{'}$
\FOR{$j=1$ to $length(D^{'})$}
\STATE $generatedSkillset \gets \{\}$
\FOR{$i=1$ to $length(cluster)$}
\STATE $count \gets D^{'}[j][Cul_{i}]$
\STATE $groupedSkills \gets mapper[Cul_{i}][0]$
\STATE $membership \gets mapper[Cul_{i}][1]$
\IF{$count$}
\IF{$count \leq length(groupedSkills)$}
\STATE Select $count$ number of skills from  $groupedSkills$ using probability distribution by $membership$: $skills^{'} \gets \text{selectSkills}(count, groupedSkills,membership)$
\ELSE
\STATE $skills^{'} \gets \text{selectSkills}(length(groupedSkills), groupedSkills,membership)$
\ENDIF
\STATE $generatedSkillset \gets generetedSkillset \bigcup \{skills^{'}\}$
\ENDIF
\ENDFOR
\STATE $D^{'}[j][synthetic\_skill] \gets generatedSkillset$
\ENDFOR
\STATE Remove all the columns from $D^{'}$ that were cluster IDs
\STATE Rename $synthetic\_skill$ to $skills$
\STATE Return $D^{'}$
\STATE End
\end{algorithmic}
\end{algorithm}

% \vspace{-0.2in}

\begin{table}[htpb]
\centering
\caption{ Frequency of unique skillsets in the example illustrated in Table \ref{example}}
\label{entropy}
\resizebox{0.8\columnwidth}{!}{%
\begin{tabular}{|c|c|c|}
\hline
\textbf{Skillsets} & \textbf{Frequency} & \textbf{\begin{tabular}[c]{@{}c@{}}Normalised\\ Frequency\end{tabular}} \\ \hline
C++, C, Java          & 1  & 0.0625 \\ \hline
HTML, Javascript      & 2  & 0.125  \\ \hline
Java, Javascript,HTML & 3  & 0.1875 \\ \hline
PHP,Javascript,HTML   & 4  & 0.25   \\ \hline
Java, PHP,Node.js     & 4  & 0.25   \\ \hline
Python,R              & 2  & 0.125  \\ \hline
\textbf{Total}        & 16 & 1      \\ \hline
\end{tabular}%
}
\end{table}

\begin{figure}[htbp]
  \centering
  \begin{minipage}[t]{0.45\textwidth}
    \centering
    \begin{minipage}[t]{0.48\textwidth}
      \centering
      \includegraphics[width=\textwidth]{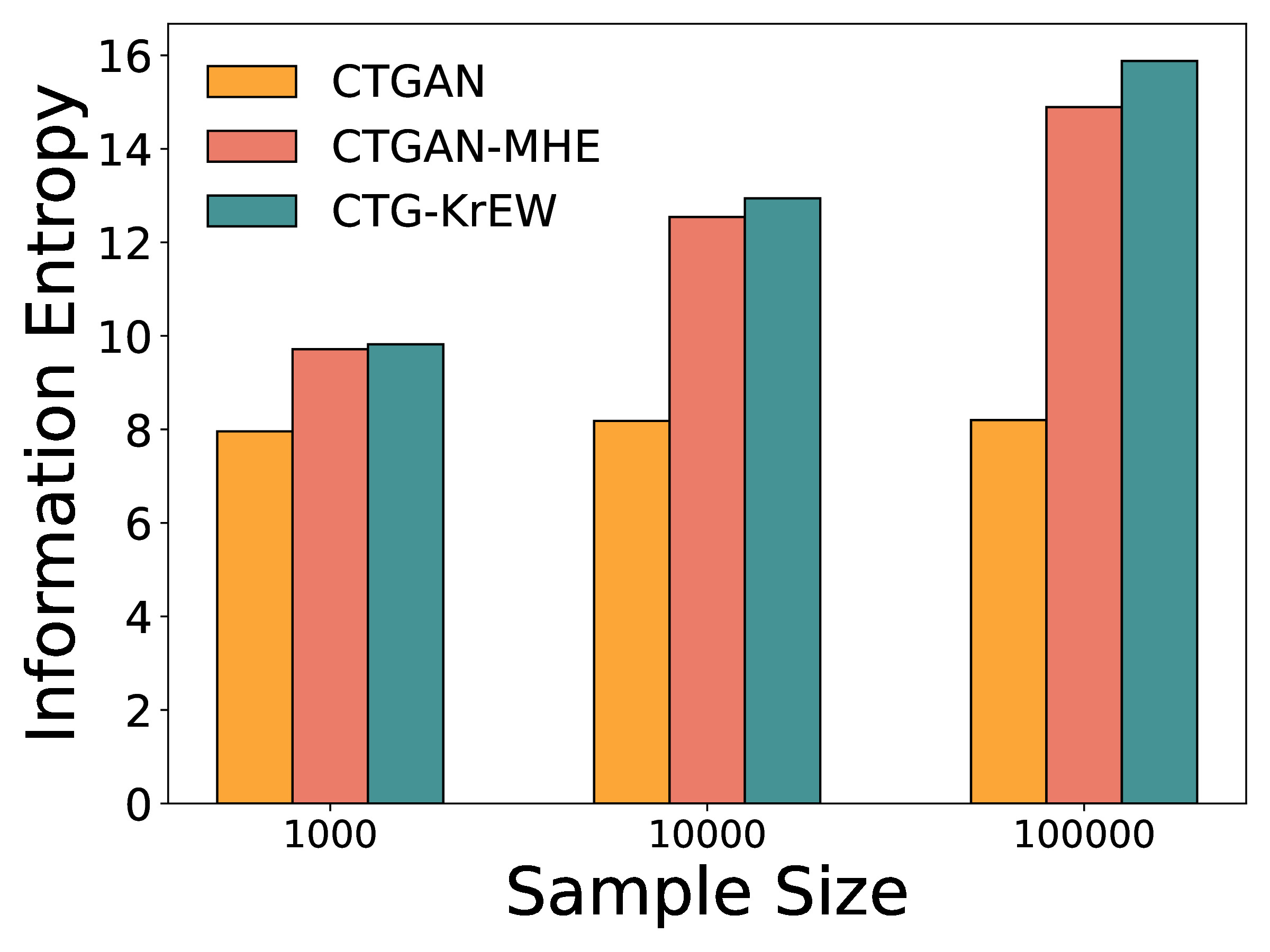}
      \subcaption{task-data}
      \label{entropy-t}
    \end{minipage}
    % \hfill
    \begin{minipage}[t]{0.48\textwidth}
      \centering
      \includegraphics[width=\textwidth]{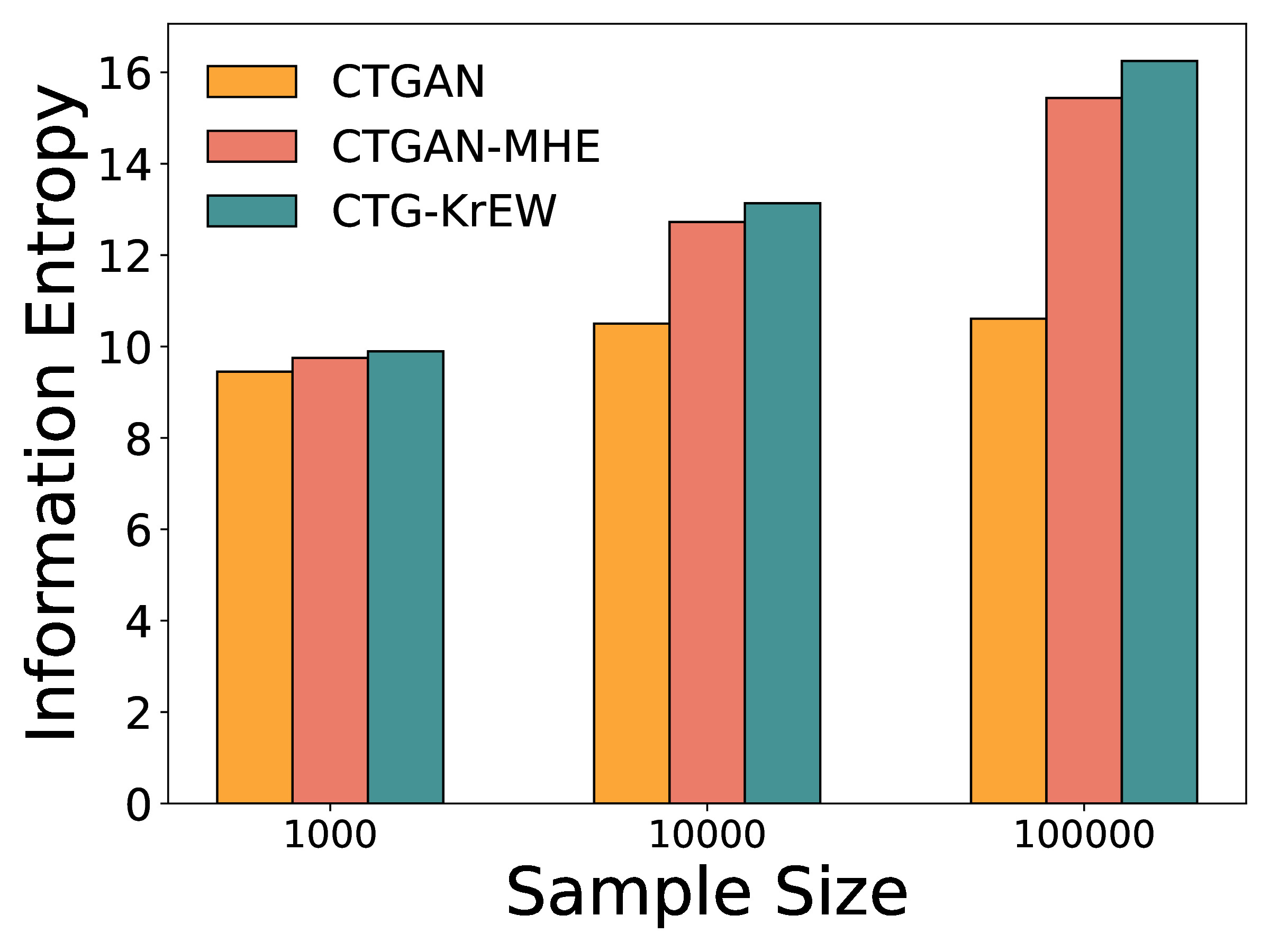}
      \subcaption{worker-data}
      \label{entropy-w}
    \end{minipage}
     \caption{Information entropy to capture the skillset variability in the generated synthetic datasets.}
  \end{minipage}
  \label{ent}
\end{figure}

The present investigation concerns the calculation of entropy for the synthetic datasets produced by each of the three variants. A significant limitation of the generic CTGAN model, when employed to generate skillsets or collections of associated skills is the introduction of repetitive structured skillsets. As a result, the information entropy, as shown in Figures \ref{entropy-t} and \ref{entropy-w}, is the lowest.  It is noteworthy that since the source \textit{task-data} consisted of only 297 unique skillsets, even when generating datasets of size 10k, only 297 distinct combinations of skillsets could be produced, leading to a significant repetition of skillsets. In contrast, the other two methods performed much better in terms of skillset variability for both \textit{task-data} and \textit{worker-data}. As a consequence of this rationale, we have concentrated our analysis exclusively on the comparative efficacy of the CTGAN model with multi-hot encoding (abbreviated as CTGAN-MHE) and \textit{CTG-KrEW}

\vspace{0.1in}
\noindent(ii) \textbf{Skillset matching:} The introduction of the \textit{skillset matching} parameter aims to effectively capture the contextual similarity among skills within a given skillset while considering the interrelationships between these skills and the overall structure of the skillset as a cohesive unit. The study employed the utilisation of a \textbf{\textit{Sentence Transformer}} \cite{ravichandiran2021getting}, a specific model of natural language processing (NLP) that encodes sentences into vectors or embeddings of fixed dimensions, thereby capturing the semantic meaning of the sentences. 

We employed a pre-trained BERT-based sentence embedding model from the Sentence-Transformer library \cite{sentence-transformers} to calculate the \textbf{\textit{cosine similarity score}} between the skillsets. The process involves comparing the similarity of each encoded skillset record in the synthetic dataset with every skillset record in the source dataset and selecting the highest similarity value as skillset matching score. As an illustration, let us consider two skillsets extracted from the synthetic dataset, denoted as $S^{'}_{1}=$ (Java, C++) and $S^{'}_{2}=$ (PHP, C++).

\begin{figure}[htbp]
  \centering
  \begin{minipage}[t]{0.45\textwidth}
    \centering
    \begin{minipage}[t]{0.48\textwidth}
      \centering
      \includegraphics[width=\textwidth]{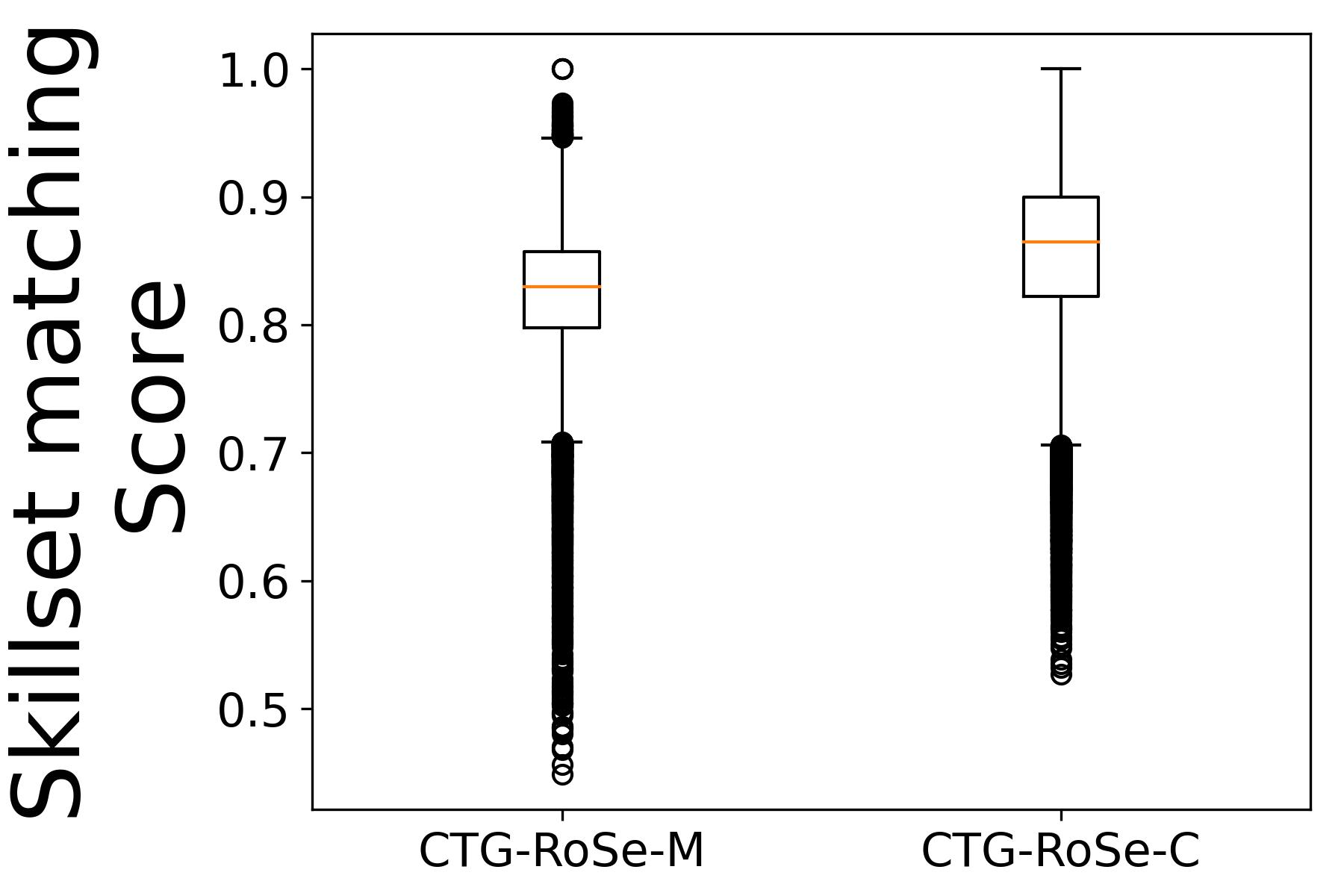}
      \subcaption{task-data}
     \label{skill_match-t}
    \end{minipage}
    % \hfill
    \begin{minipage}[t]{0.48\textwidth}
      \centering
      \includegraphics[width=\textwidth]{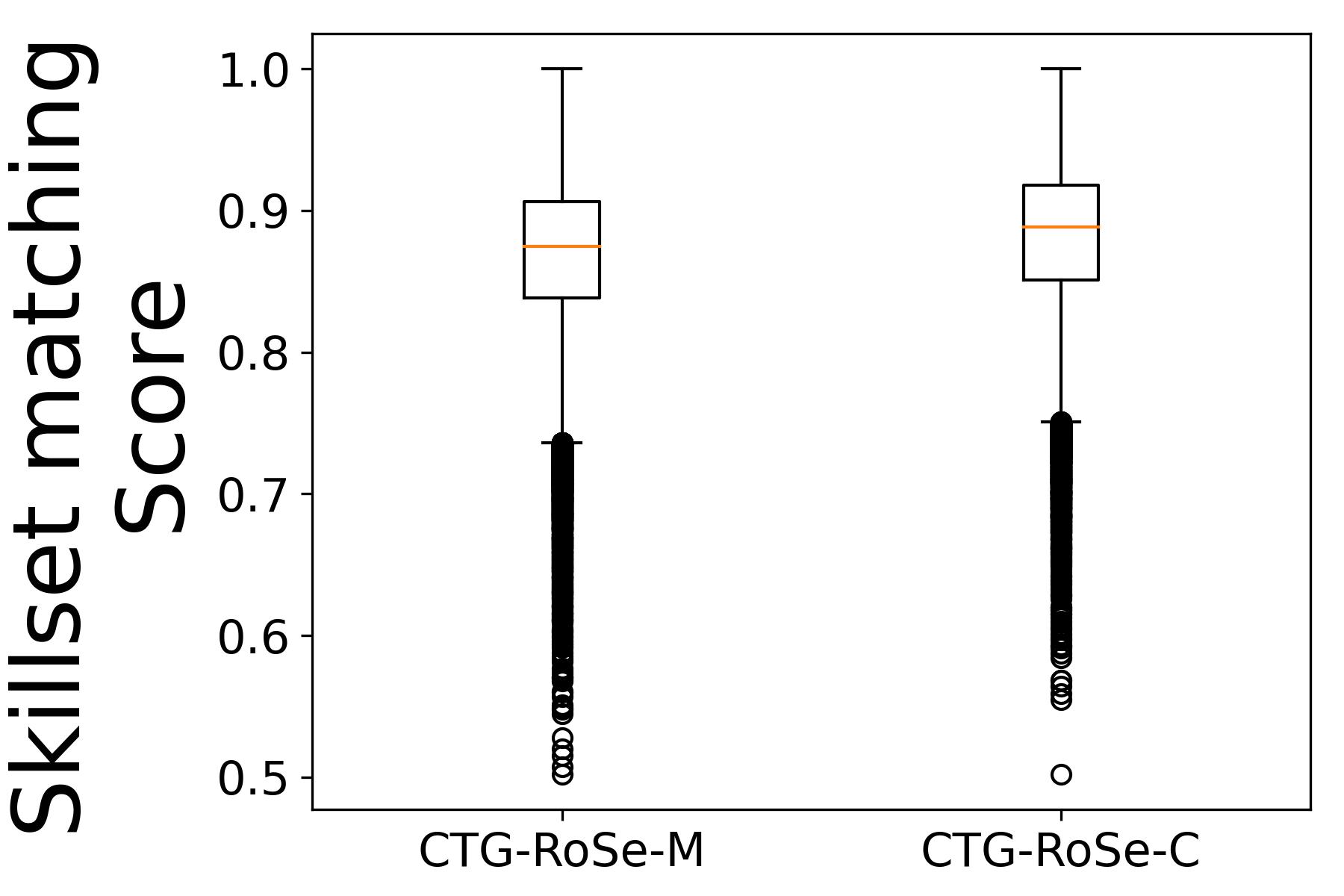}
      \subcaption{worker-data}
     \label{skill_match-w}
    \end{minipage}
     \caption{Skillset matching score.}
  \end{minipage}
   \label{skill_match}
\end{figure}

Upon computing the similarity scores between $S^{'}_{1}$ and $S^{'}_{2}$ with the 7 skillset records presented in Table \ref{example}, we obtain the following values: [0.985,0.847,0.845 ,0.817,0.853 ,0.568,0.847] and [0.896, 0.776, 0.761, 0.840, 0.874, 0.642, 0.776], respectively. It is observed that the maximum value for $S^{'}_{1}$ is 0.985, while the maximum value for $S^{'}_{2}$ is 0.896. These results indicate that (Java, C++) exhibits a higher degree of semantic similarity to (C++, C, Java) compared to (PHP, C++). Drawing upon this concept, we endeavoured to capture the contextual similarity between the skillsets of the source and synthetic domains. The mean value of all the scores is utilised for comparison. Box plots illustrate depicted in Figures \ref{skill_match-t} and \ref{skill_match-w}. indicate that the average skillset matching score of CTGAN-MHE exhibited superior performance compared to \textit{CTG-KrEW} about both the \textit{task-data} and \textit{worker-data}.

\noindent(iii) \textbf{Similarity of the skills distribution:} To assess the likelihood that the skills present in the synthetic dataset exhibit a frequency distribution that closely aligns with the skills in the original dataset, \textit{\textbf{KL divergence metric}} \cite{kullback1951information} is used. KL or Kullback–Leibler divergence metric, denoted as $D_{KL}(P||Q)$, is a statistical measure that represents a quantification of the dissimilarity between a given probability distribution $P$ and a reference probability distribution $Q$. 

To compute the KL divergence of the dataset, we adopt a methodology similar to that employed for assessing the variability of the skillset. However, instead of determining the frequency of distinct skillsets, we ascertain the frequency of unique skills across all the skillsets. As demonstrated in Table \ref{example}, the number of distinct skillsets was $m=6$, while the count of the number of unique skills was $n=9$ and thus generated Table \ref{freq} displays the frequency of the nine skills.

\begin{table}[htpb]
\centering
\caption{ Frequency of skills in the example illustrated in Table \ref{example} and its assumed synthetic equivalent.}
\label{freq}
\renewcommand{\arraystretch}{1.2}
\setlength{\tabcolsep}{4pt}
\resizebox{\columnwidth}{!}{%
\begin{tabular}{|c|cc|cc|}
\hline
\multirow{2}{*}{\textbf{Skill}} &
  \multicolumn{2}{c|}{\textbf{Frequency}} &
  \multicolumn{2}{c|}{\textbf{\begin{tabular}[c]{@{}c@{}}Normalised \\ Frequency\end{tabular}}} \\ \cline{2-5} 
 &
  \multicolumn{1}{c|}{\textbf{Source dataset}} &
  \textbf{\begin{tabular}[c]{@{}c@{}}Synthetic dataset\\ (Assumed)\end{tabular}} &
  \multicolumn{1}{c|}{\textbf{\begin{tabular}[c]{@{}c@{}}Source dataset\\ (freq/18)\end{tabular}}} &
  \textbf{\begin{tabular}[c]{@{}c@{}}Synthetic dataset\\ (freq/16)\end{tabular}} \\ \hline
C++            & \multicolumn{1}{c|}{1}  & 2  & \multicolumn{1}{c|}{0.05555555556} & 0.125  \\ \hline
C              & \multicolumn{1}{c|}{1}  & 1  & \multicolumn{1}{c|}{0.05555555556} & 0.0625 \\ \hline
Java           & \multicolumn{1}{c|}{3}  & 2  & \multicolumn{1}{c|}{0.1666666667}  & 0.125  \\ \hline
Html           & \multicolumn{1}{c|}{4}  & 4  & \multicolumn{1}{c|}{0.2222222222}  & 0.25   \\ \hline
Js             & \multicolumn{1}{c|}{4}  & 2  & \multicolumn{1}{c|}{0.2222222222}  & 0.125  \\ \hline
Php            & \multicolumn{1}{c|}{2}  & 1  & \multicolumn{1}{c|}{0.1111111111}  & 0.0625 \\ \hline
Nodejs         & \multicolumn{1}{c|}{1}  & 1  & \multicolumn{1}{c|}{0.05555555556} & 0.0625 \\ \hline
python         & \multicolumn{1}{c|}{1}  & 2  & \multicolumn{1}{c|}{0.05555555556} & 0.125  \\ \hline
R              & \multicolumn{1}{c|}{1}  & 1  & \multicolumn{1}{c|}{0.05555555556} & 0.0625 \\ \hline
\textbf{Total} & \multicolumn{1}{c|}{18} & 16 & \multicolumn{1}{c|}{1}             & 1      \\ \hline
\end{tabular}%
}
\end{table}

\begin{table*}[htpb]
\centering
\small
\begin{blockarray}{cccccccccc}
& \text{C++} & \text{C} & \text{Java} & \text{HTML} & \text{JavaScript} & \text{PHP} & \text{Node.js} & \text{Python} & \text{R}\\
\begin{block}{c(ccccccccc)}
\text{C++} & 0 & 1 & 1 & 0 & 0 & 0 & 0 & 0 & 0 \\
\text{C} & 1 & 0 & 1 & 0 & 0 & 0 & 0 & 0 & 0\\
\text{Java} & 1 & 1 & 0 & 1 & 1 & 1 & 1 & 0 &0 \\
\text{HTML} & 0 & 0 & 1 & 0 & 3 & 1 & 0 & 0 &0\\
\text{JavaScript} & 0 & 0 & 1 & 3 & 0 & 1 & 0 & 0 &0\\
\text{PHP} & 0 & 0 & 1 & 1 & 1 & 0 & 1 & 0 &0\\
\text{Node.js} & 0 & 0 & 1 & 0 & 0 & 1 & 0 & 0 &0\\
\text{Python} & 0 & 0 & 0 & 0 & 0 & 0 & 0 & 0 &1\\
\text{R} & 0 & 0 & 0 & 0 & 0 & 0 & 0 & 1 &0\\
\end{block}
\end{blockarray}
\caption{Association matrix of the skills from Table \ref{example}}
\label{asso}
\end{table*}
\vspace{-0.1in}

\vspace{0.1in}
Therefore, by referring to Table \ref{freq}, we obtain two distributions: $P$, representing the normalized frequency of the source dataset, and $Q$, representing the normalized frequency of the synthetic dataset. Subsequently, we calculate $D_{KL}(P || Q)$ using the following formula: 
\begin{equation*}
{\displaystyle D_{\text{KL}}(P\parallel Q)=\sum _{x\in {\mathcal {X}}}P(x)\log \left({\frac {P(x)}{Q(x)}}\right)}
\end{equation*}
where $\mathcal{X}$ denotes the collection of all conceivable values of the random variable under examination. The unbounded nature of the KL divergence signifies that it can assume any non-negative value or even tend towards infinity. However, a KL divergence value of zero indicates that the two distributions being compared possess equivalent amounts of information.

\begin{figure}[htbp]
  \centering
  \begin{minipage}[t]{0.45\textwidth}
    \centering
    \begin{minipage}[t]{0.48\textwidth}
      \centering
      \includegraphics[width=\textwidth]{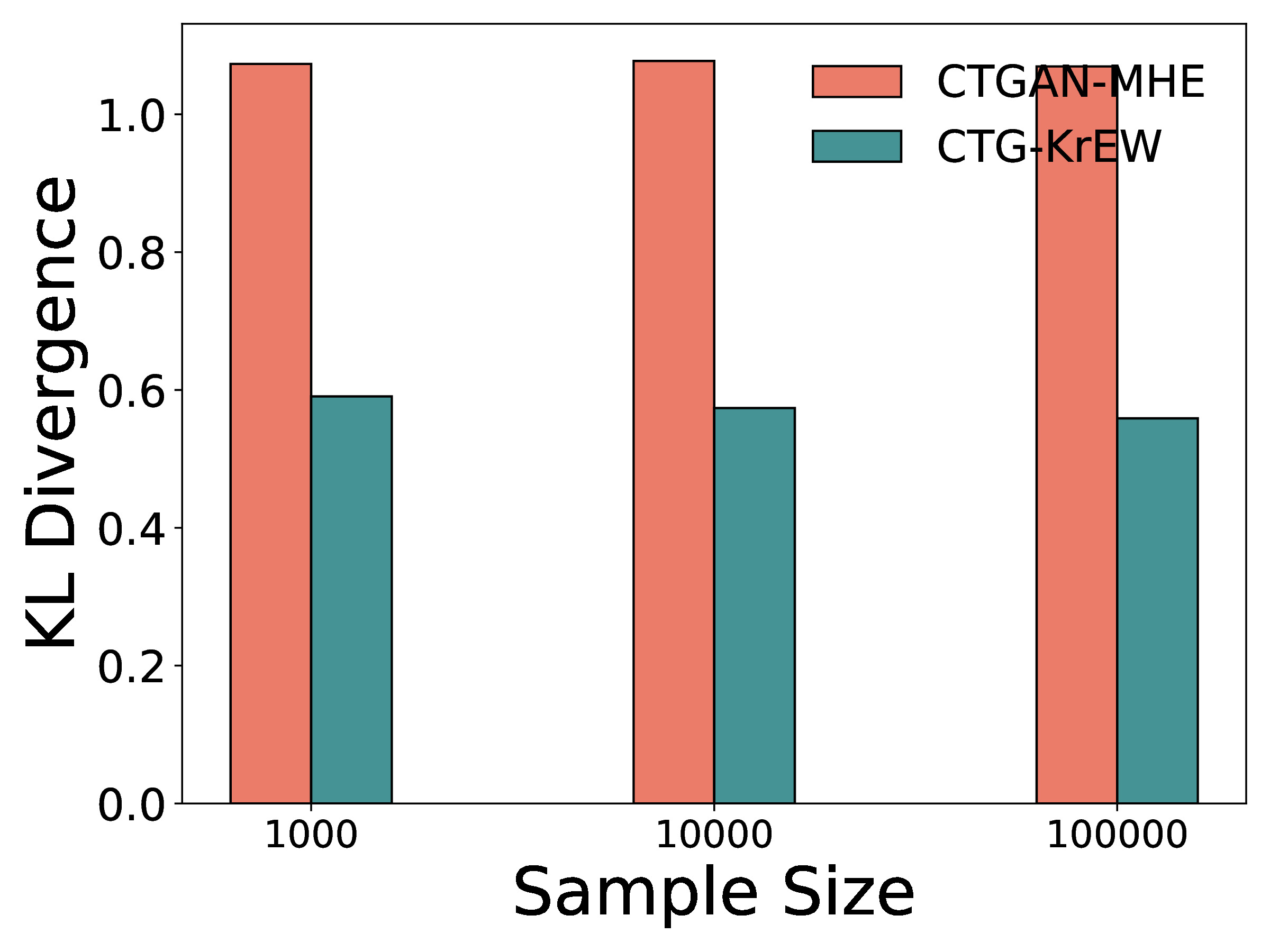}

      \subcaption{task-data}
      \label{kl-t}
    \end{minipage}
    % \hfill
    \begin{minipage}[t]{0.48\textwidth}
      \centering
      \includegraphics[width=\textwidth]{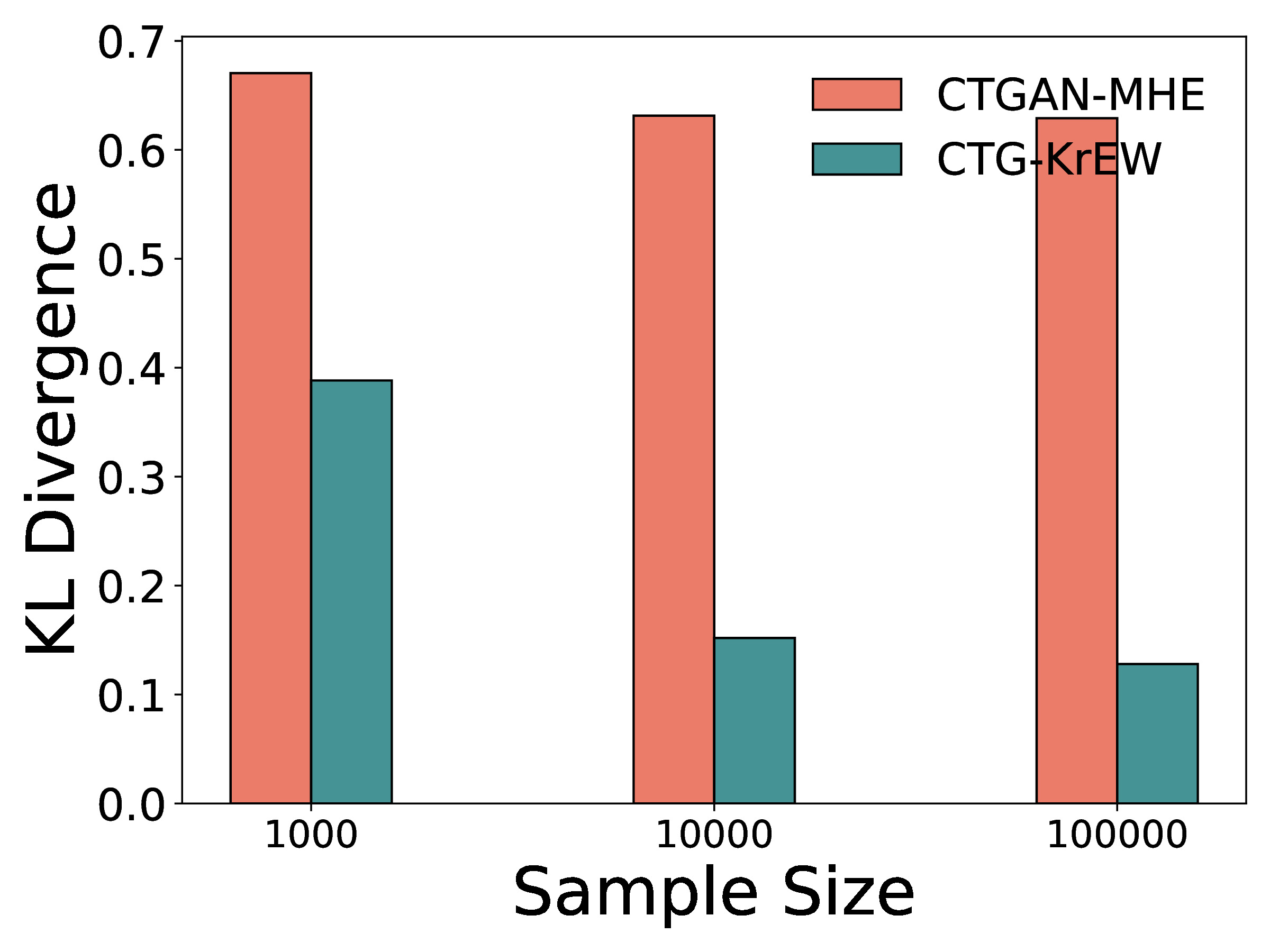}

      \subcaption{worker-data}
      \label{kl-w}
    \end{minipage}
     \caption{KL divergence score to compare the similarity of the `skill' distribution between the source and synthetic dataset.}
  \end{minipage}
  \label{kl}
  \vspace{-0.1in}
\end{figure}

Figures \ref{kl-t} and \ref{kl-w} indicate that the KL divergence score is lowest for \textit{CTG-KrEW} in comparison to  CTGAN-MHE which suggests that the synthetic data generated by \textit{CTG-KrEW} exhibits the highest degree of similarity to the frequency distribution of the `skill' attribute in the original dataset.  This similarity applies to both the categories of \textit{task-data} and \textit{worker-data}. Moreover, as the size of the sample data increases, the KL divergence score decreases specifically for the \textit{worker-data}. The observed incongruity can be attributed to the fact that the initial size of the \textit{worker-data} was notably larger than that of the \textit{task-data}, leading to increased diversity among the distinct skills (as well as other attributes).

\vspace{0.1in}
\noindent(iv) \textbf{Associativity among the skills:} Maintaining the associativity between the skills presented posed a persistent challenge that remained unresolved by the generic CTGAN and CTGAN-MHE approaches. The development of \textit{CTG-KrEW} was significantly influenced by this particular challenge.

To ensure that the skillsets produced are consistent with practical requirements and adequately maintain the joint distribution of the `skill' column with other attributes in the synthetic dataset, it is imperative to remain attentive to the associativity or co-occurrence of individual skills within their respective skill sets. Consequently, the \textit{\textbf{Pearson correlation coefficient}} \cite{glen2021correlation} metric is used to capture the associativity between skills that co-occur within skillsets. To find the Pearson's correlation coefficient ($\rho$) in our datasets, both source and synthetic, an association matrix of size $(n \times n)$ is first built, where $n$ is the number of unique skills in the data set under consideration. For example, from Table \ref{example}, the association matrix (see Table \ref{asso}) can be formed.

In Table \ref{asso}, the value in cell $(i,j)$ represents the number of times that skill $i$ co-occurs with skill $j$ in all skill sets and vice versa. If skill $i$ never cooccurs with skill $j$, then cell $(i,j)$ contains zero. This $(9 \times 9)$ matrix is then normalized and transformed into a $(1 \times 81)$ vector. A similar vector is created for the skills in the synthetic dataset. For a pair of random variables $(X,Y)$, the formula for the Pearson correlation coefficient is given by:
\begin{equation*}
\rho(X,Y) = \frac{\operatorname{cov}(X,Y)}{\sigma_X \sigma_Y}
\end{equation*}
where $\operatorname{cov}$ denotes the covariance, $\sigma_X$ is the standard deviation of $X$, and $\sigma_Y$ is the standard deviation of $Y$. In this context, $X$ and $Y$ correspond to the associative vectors of the source and synthetic datasets, respectively.

\begin{figure}[htbp]
  \centering
  \begin{minipage}[t]{0.45\textwidth}
    \centering
    \begin{minipage}[t]{0.48\textwidth}
      \centering
      \includegraphics[width=\textwidth]{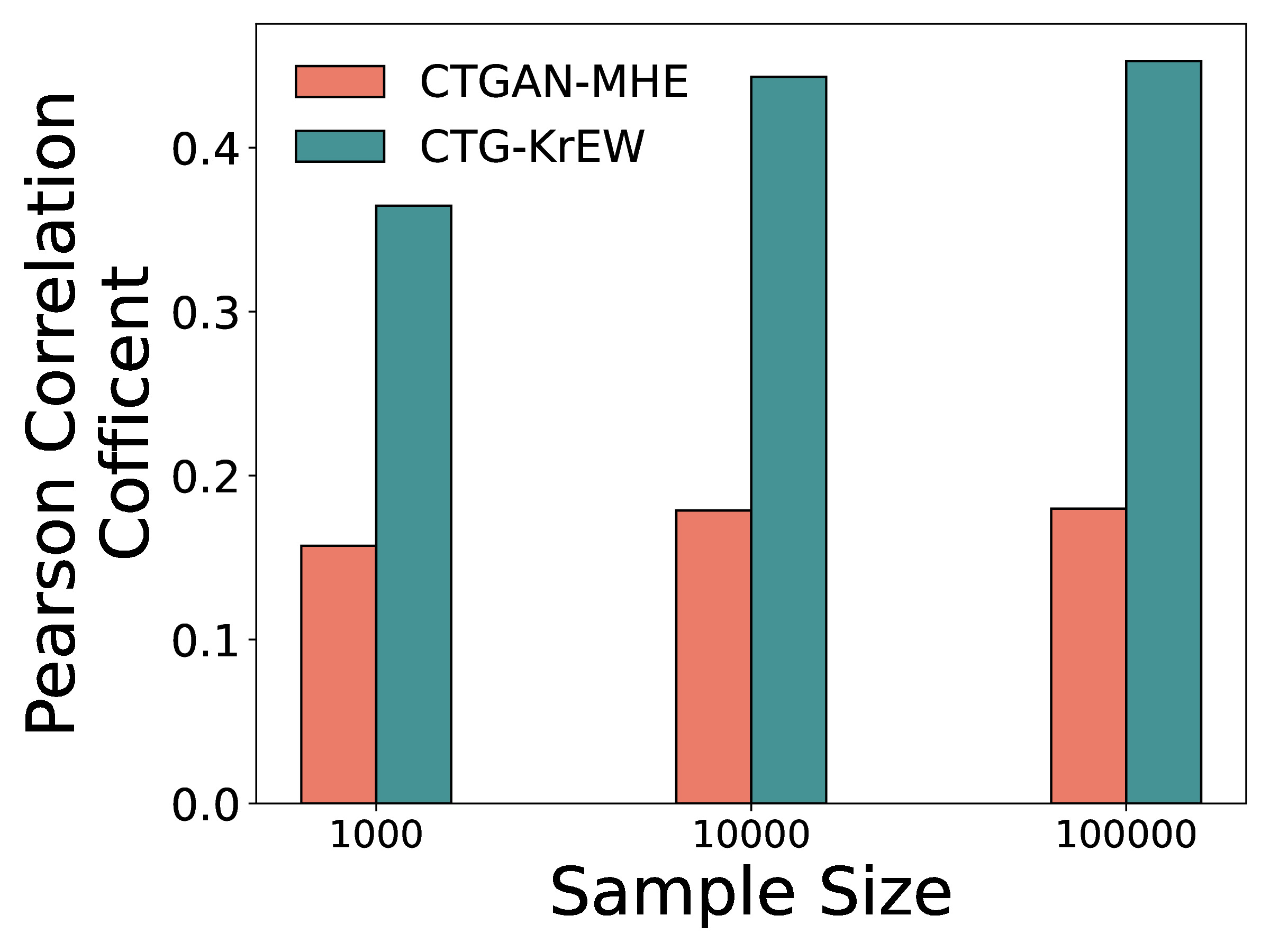}
      \subcaption{task-data}
      \label{rho-t}
    \end{minipage}
    % \hfill
    \begin{minipage}[t]{0.48\textwidth}
      \centering
      \includegraphics[width=\textwidth]{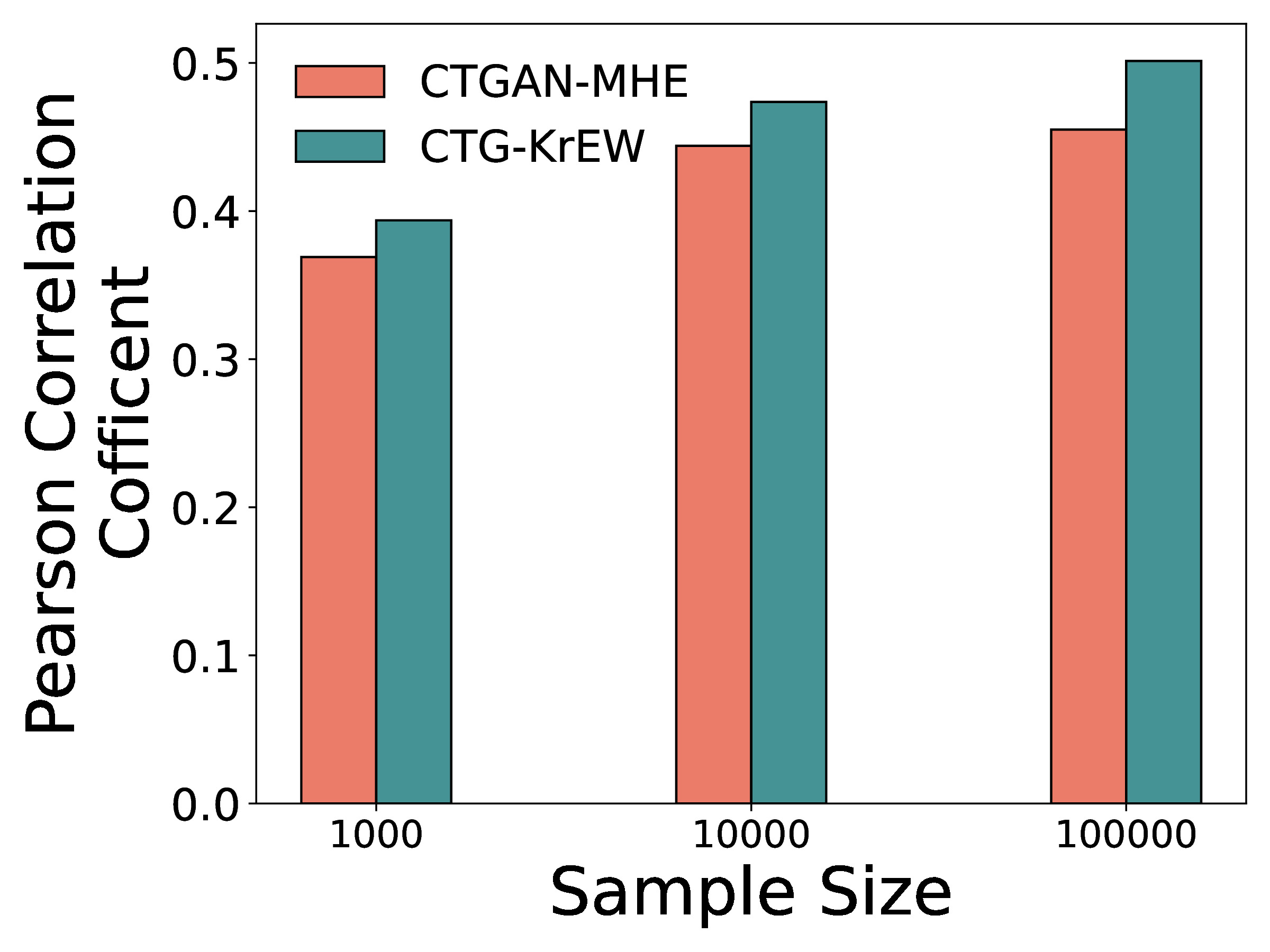}
      \subcaption{worker-data}
      \label{rho-w}
    \end{minipage}
     \caption{Pearson correlation coefficient score to compare the associativity among the skills distribution between the source and synthetic dataset. }
  \end{minipage}
  \label{rho}
\end{figure}

The results depicted in Figures \ref{rho-t} and \ref{rho-w} indicate that for \textit{CTG-KrEW}, there is a greater degree of similarity in associativity among the individual skills within their respective skillsets when comparing the source and synthetic datasets, concerning CTGAN-MHE.

\vspace{-0.1in}

\vspace{0.1in}
\noindent(v) \textbf{System feasibility:}For this analysis, we conducted a comparison of the variants based on their \textbf{\textit{memory usage}} and \textbf{\textit{training time}} consumption. It should be noted that after the training and deployment of the models, the process of generating a dataset comprising even 10,000 records is relatively quick, regardless of the variant utilized.

\begin{figure}[!t]
  \centering
  \begin{minipage}[t]{0.45\textwidth}
    \centering
    \begin{minipage}[t]{0.48\textwidth}
      \centering
      \includegraphics[width=\textwidth]{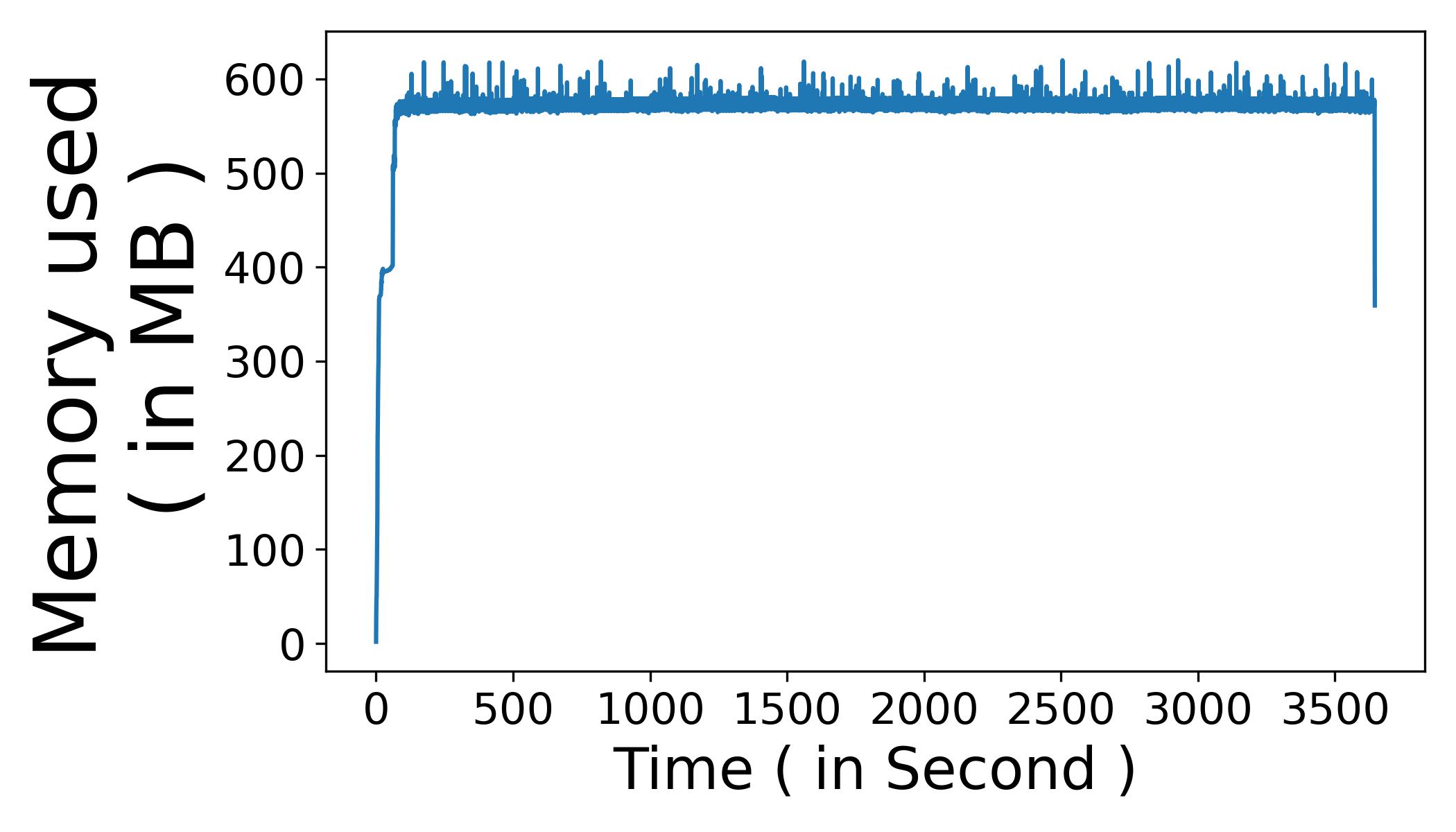}
      \subcaption{CTGAN-MHE}
    \end{minipage}
    \hfill
    \begin{minipage}[t]{0.48\textwidth}
      \centering
      \includegraphics[width=\textwidth]{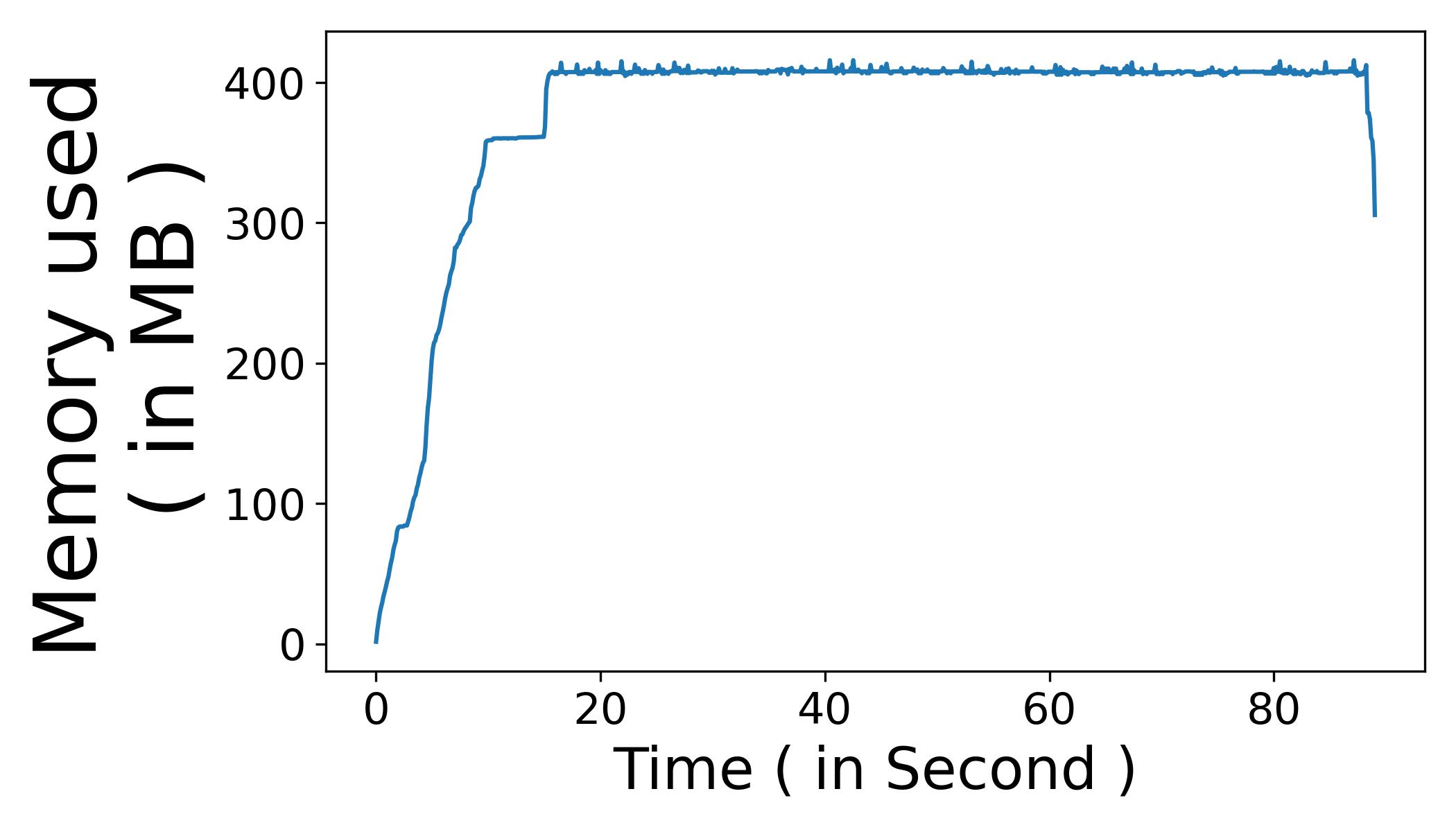}
      \subcaption{\textit{CTG-KrEW}}
    \end{minipage}
   \caption{Memory usage and training time required by the variants for task-data.}
    \label{mem_M}
  \end{minipage}
  \begin{minipage}[t]{0.45\textwidth}
    \centering
    \begin{minipage}[t]{0.48\textwidth}
      \centering
      \includegraphics[width=\textwidth]{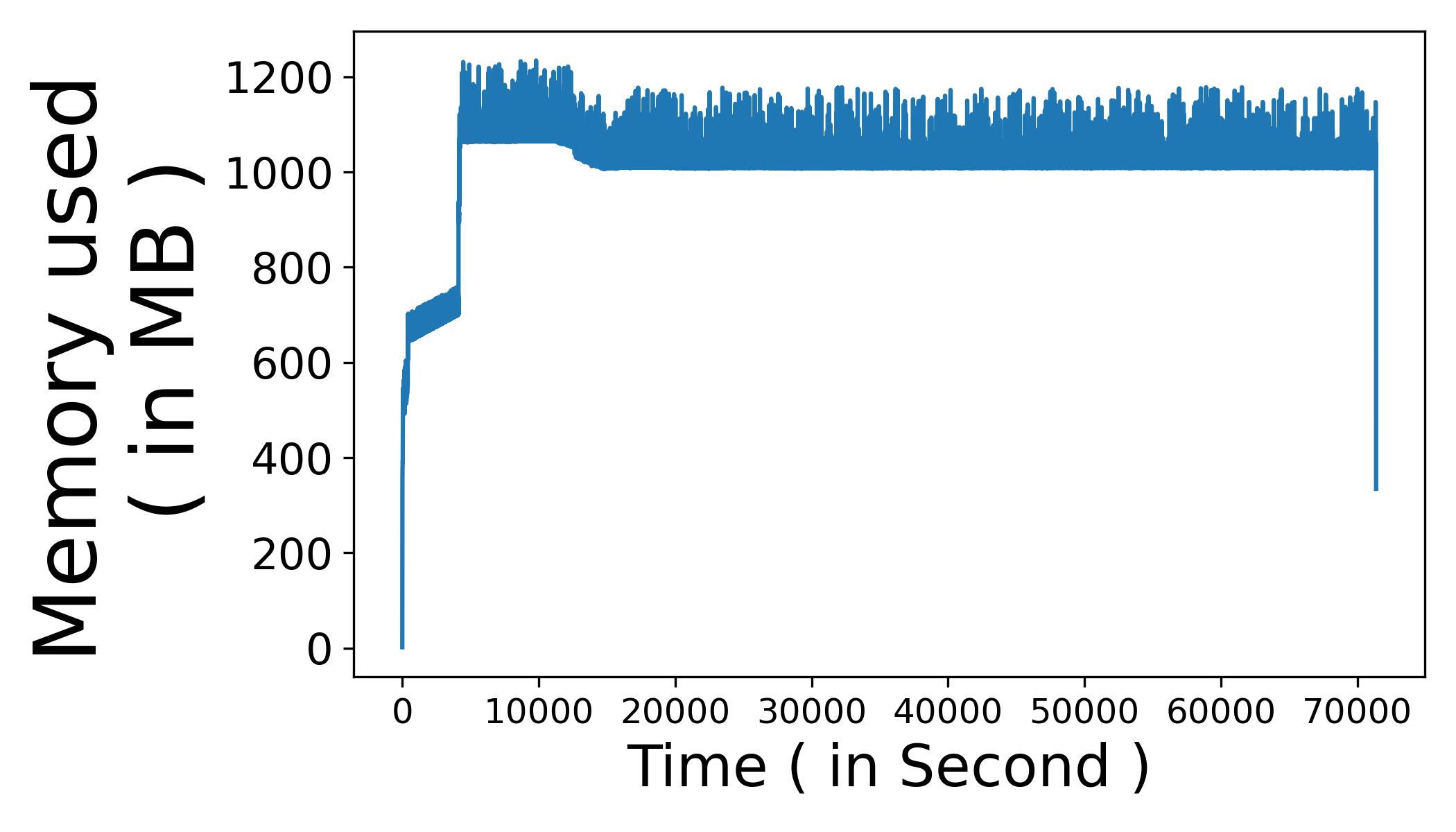}
      \subcaption{CTGAN-MHE}
    \end{minipage}
    \hfill
    \begin{minipage}[t]{0.48\textwidth}
      \centering
      \includegraphics[width=\textwidth]{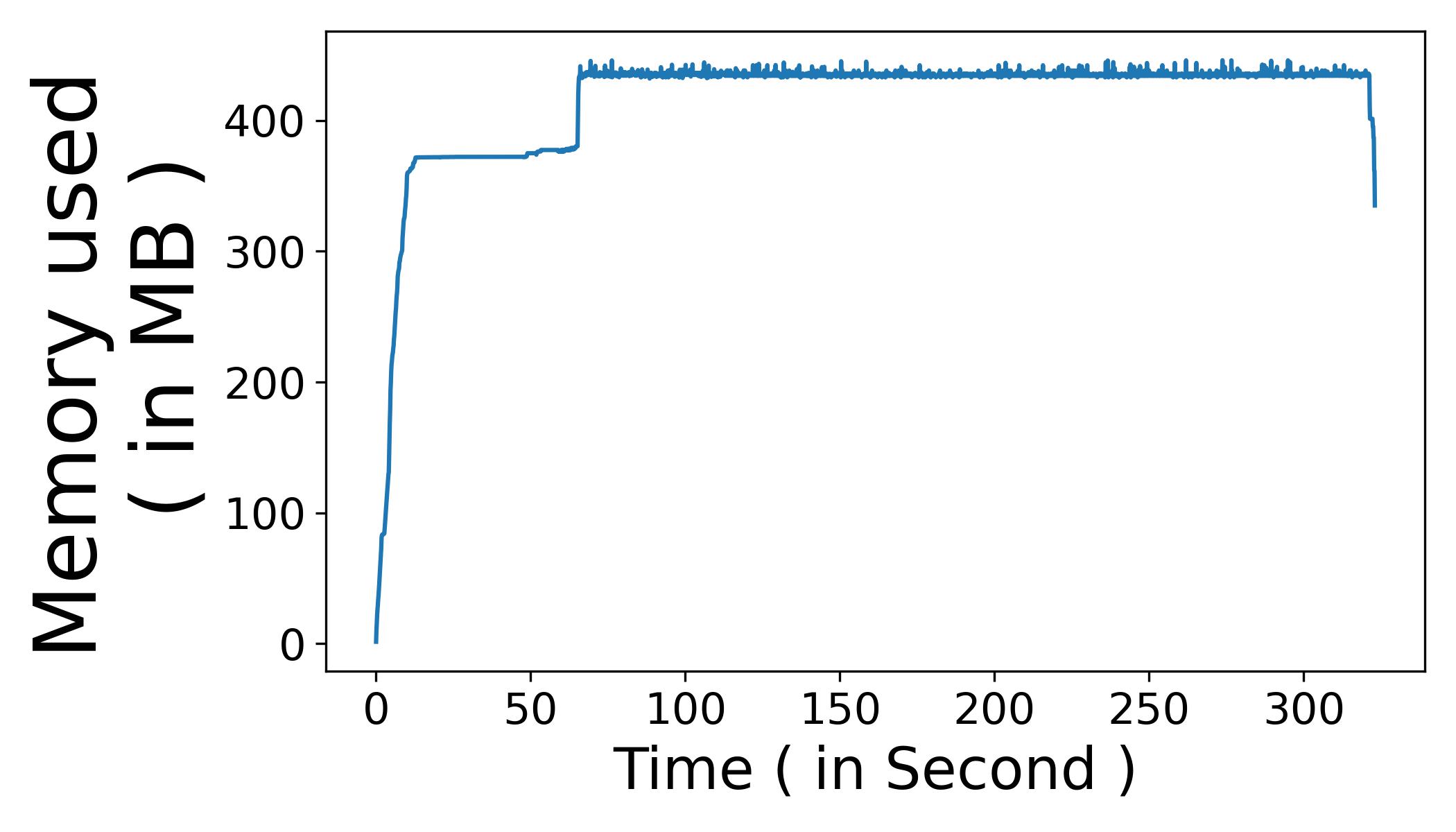}
     \subcaption{\textit{CTG-KrEW}}
    \end{minipage}
      \caption{Memory usage and training time required by the variants for worker-data.}
      \label{mem_C}
  \end{minipage}
  \vspace{-0.1in}
\end{figure}

The system feasibility results are depicted in Figures \ref{mem_M} and 9, which include 350 epochs of training. The \textit{ memory profiler} library for Python was used \cite{memory-profiler}. Memory usage and training time for the different variants using \textit{ task data} are depicted in Figure \ref{mem_M}, while Figure 11 shows the corresponding metrics for \textit{ worker data}. The empirical findings indicate that the training time required for CTGAN-MHE to complete 305 epochs is more than 300X longer than that of \textit{CTG-KrEW}. Furthermore, the \textit{CTG-KrEW} model exhibits better memory consumption performance.

\noindent(vi) \textbf{Quality of the data related to the remaining attributes:}

The quality of the attributes of both \textit{task-data} and \textit{worker-data}, excluding the `skills' column, has been illustrated in Figures 12 and Figure \ref{worker_qua}, respectively. 

All other attributes are categorical, except `fixed\_price' and `success\_rate'. The frequency of occurrence for each category or value in each attribute is calculated to compare the synthetic datasets produced by the \textit{CTG-KrEW} model with the source datasets. This analysis was performed on synthetic datasets consisting of 1000 tuples. We created fixed-size bins for the values `fixed\_price' and `success\_rate' and generated frequency histograms to show the distribution of values in each bin. The frequency counts of each characteristic are normalised to the interval $[0,1]$ to offer a clearer visualisation. The illustrations show that the patterns seen in the source data closely resembled the frequency distribution of attribute values in the synthetic data produced.

\begin{figure}[htbp]
  \centering
  \begin{minipage}[t]{0.45\textwidth}
    \centering
    \begin{minipage}[t]{0.48\textwidth}
      \centering
      \includegraphics[width=\textwidth]{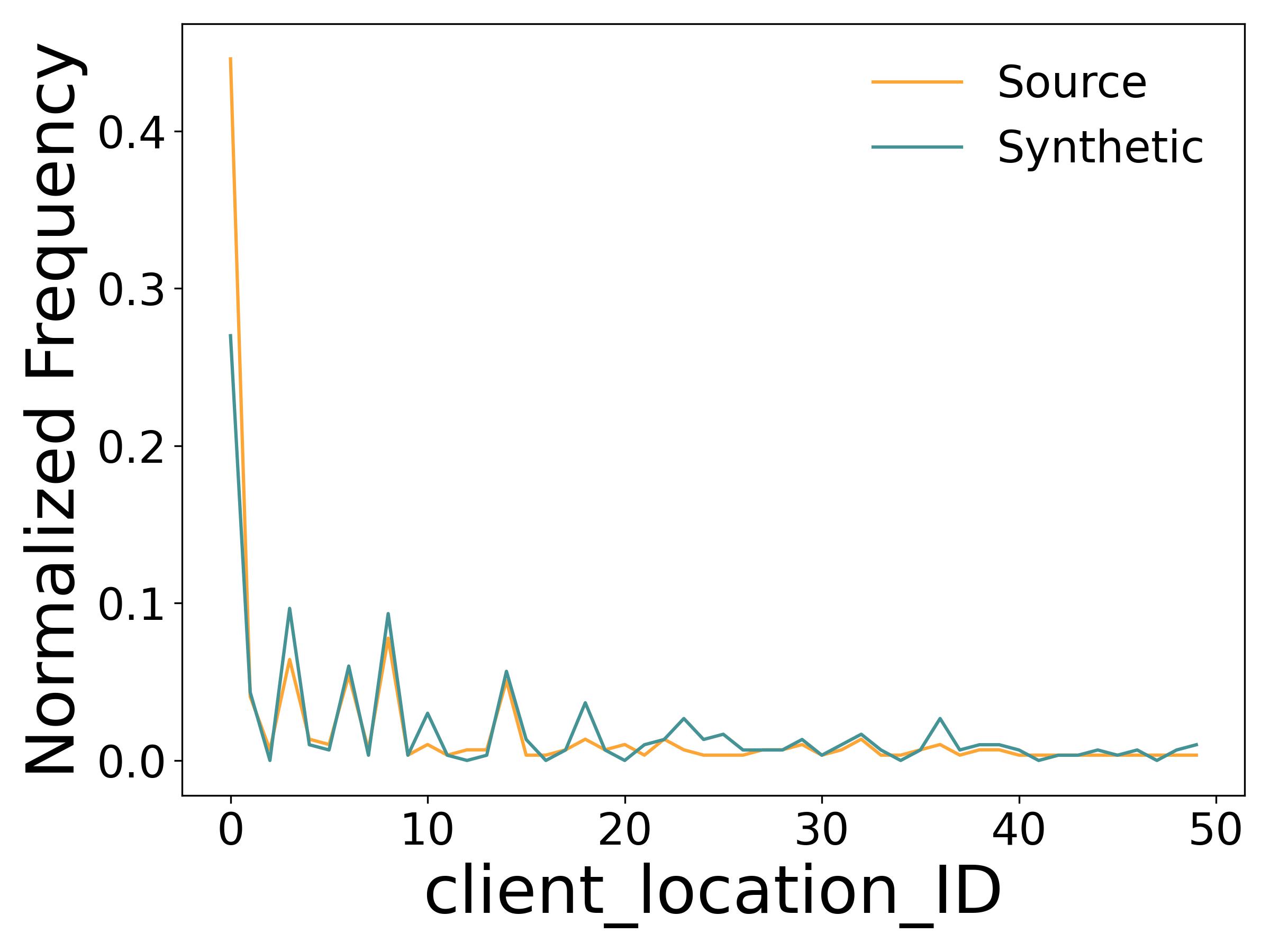}
      \subcaption{client\_location ID}
    \end{minipage}
    \hfill
    \begin{minipage}[t]{0.48\textwidth}
      \centering
      \includegraphics[width=\textwidth]{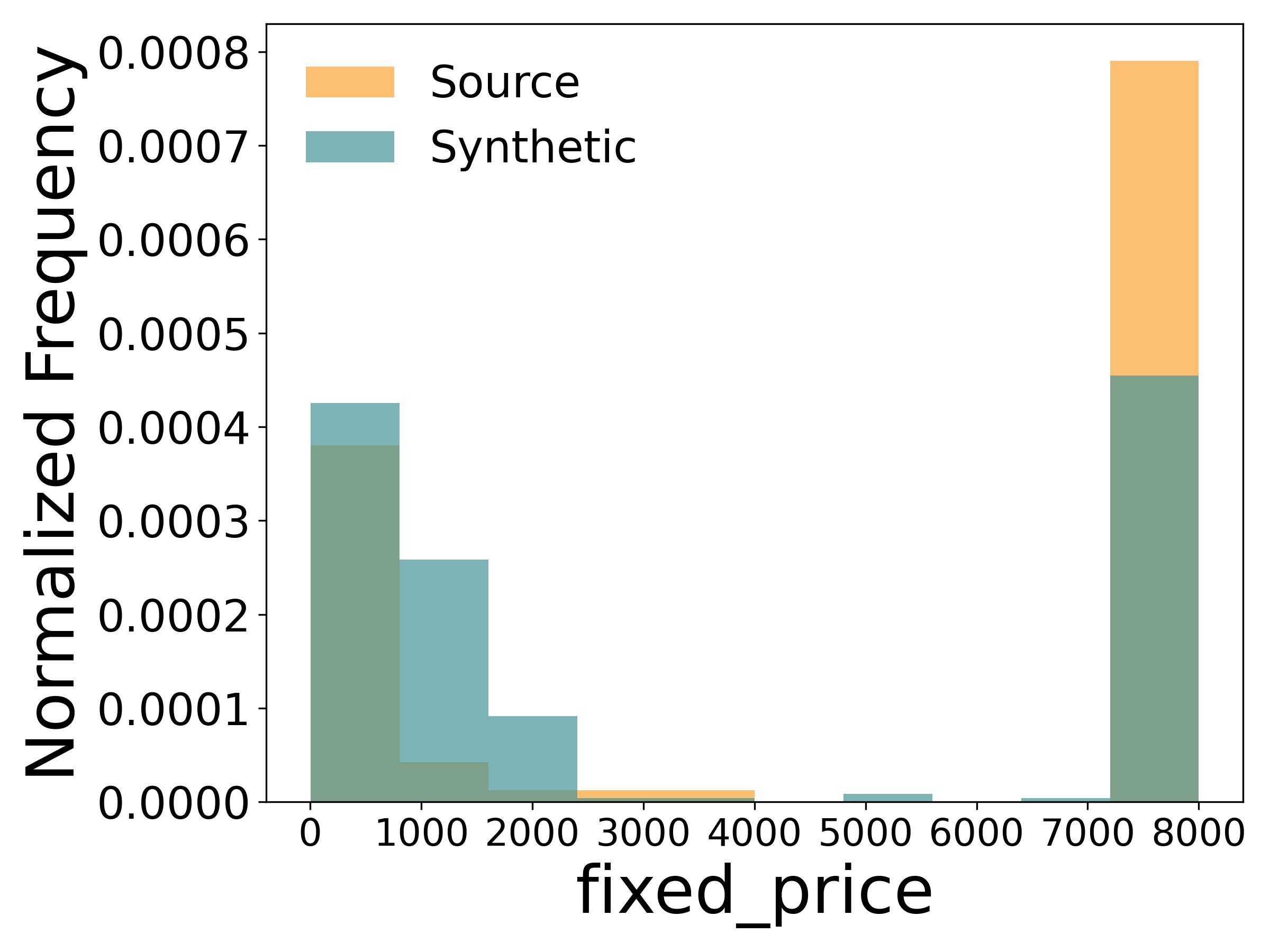}
      \subcaption{fixed\_price(budget)}
    \end{minipage}
  \end{minipage}
  \begin{minipage}[t]{0.45\textwidth}
    \centering
    \begin{minipage}[t]{0.48\textwidth}
      \centering
      \includegraphics[width=\textwidth]{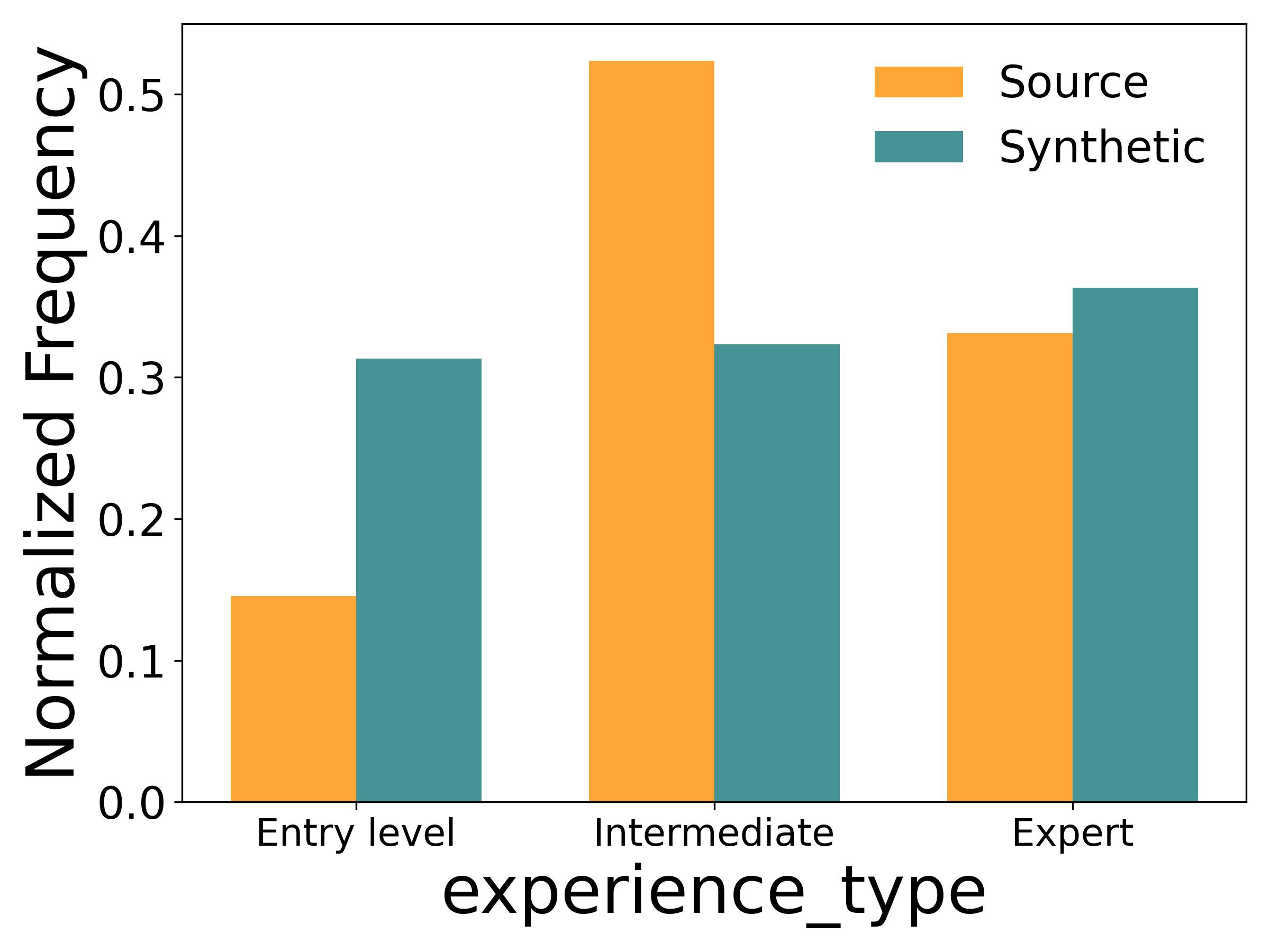}
      \subcaption{experience\_type}
    \end{minipage}
    \hfill
    \begin{minipage}[t]{0.48\textwidth}
      \centering
      \includegraphics[width=\textwidth]{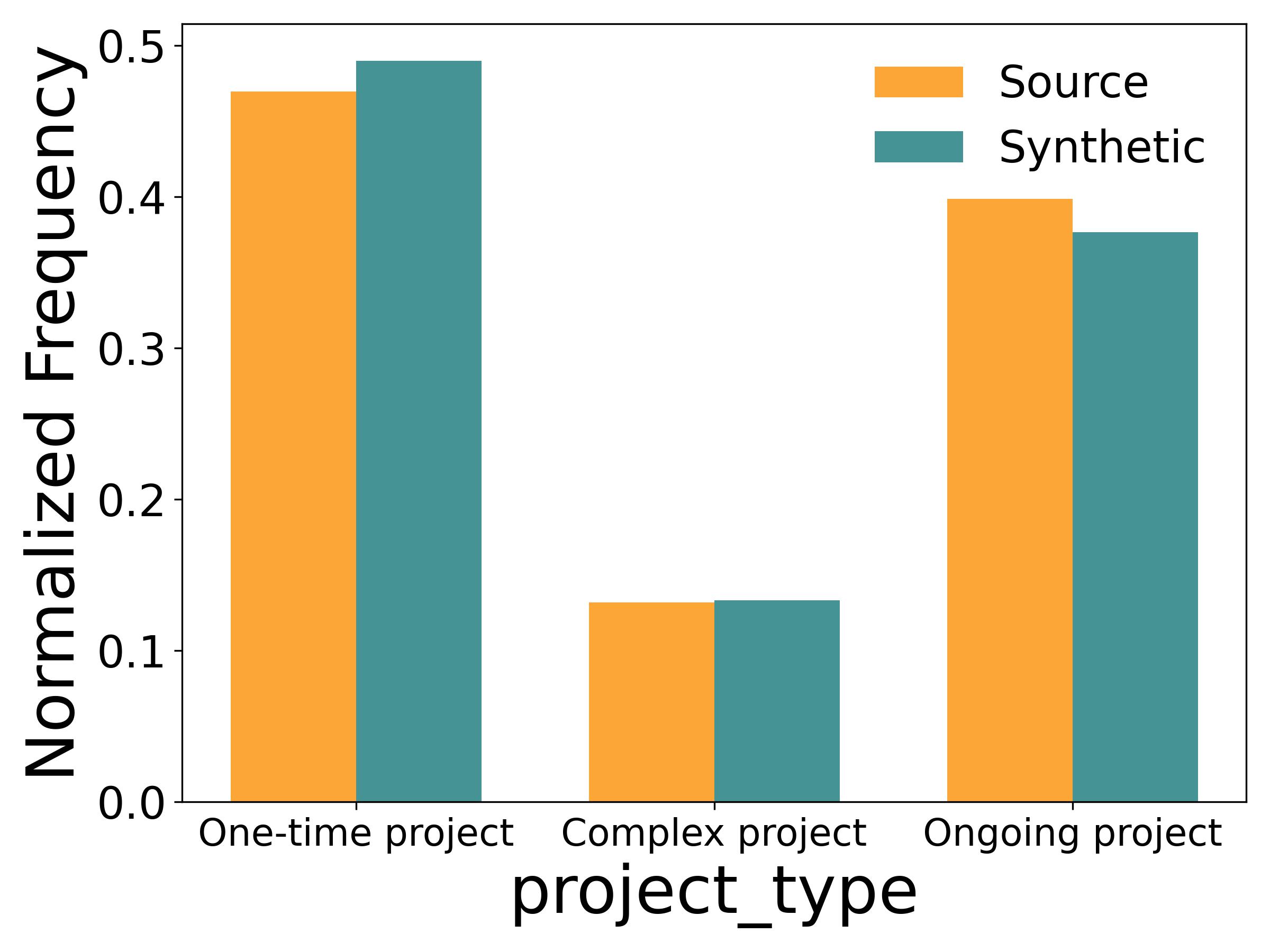}
     \subcaption{project\_type}
    \end{minipage}
      \caption{Quality of the data related to the remaining attributes of \textit{task-data}}
  \end{minipage}
   \label{task_qua}
   \vspace{-0.1in}
\end{figure}

\vspace{-0.2in}
\section{KrEW Web Application}
\label{tool}
\vspace{-0.1in}
This section provides an overview of the procedural and technical aspects of implementing and using the \textit{KrEW} application. \textit{CTG-KrEW}, recognized for its superior performance compared to baselines, utilizes its pre-trained model within the application. The model is trained on real datasets by the application's owner or administrator using \textit{CTG-KrEW}. Post-training, the model weights are stored in a pickle file, which is then loaded onto the application server. \textit{CTG-KrEW} organizes the initial dataset into distinct clusters, and training is conducted solely on attributes within these clusters. Consequently, any synthetic data generated by \textit{CTG-KrEW} follows this clustered format, necessitating a conversion back to the original raw form. The application manages these background tasks, abstracting complexities from end-users. Users interact with the application by selecting the dataset, the type of data (worker, task, or both) and the desired sample size through the application interface. Upon clicking ``Download," the request is sent to the server, where samples are generated using the pre-trained model. Upon completion, a CSV file containing the data is available for download to the user's device. Users can generate and retrieve multiple samples as needed. The application server is deployed using Flask, a Python microweb framework, with the user interface developed using HTML, CSS, and 
JavaScript. Currently, only the UpWork dataset is integrated, but the administrator can incorporate additional real datasets for training the \textit{CTG-KrEW} model. Users can choose to extract tasks-related, worker-related, or both types of information based on their requirements. For reference, Figure \ref{tool_pic} provides a snapshot of the KrEW application, accessible to users through the link \url{https://riyasamanta.github.io/krew.html}.

\begin{figure}[htpb]
  \centering
 \begin{minipage}{0.2\textwidth}
    \centering
    \includegraphics[width=\linewidth]{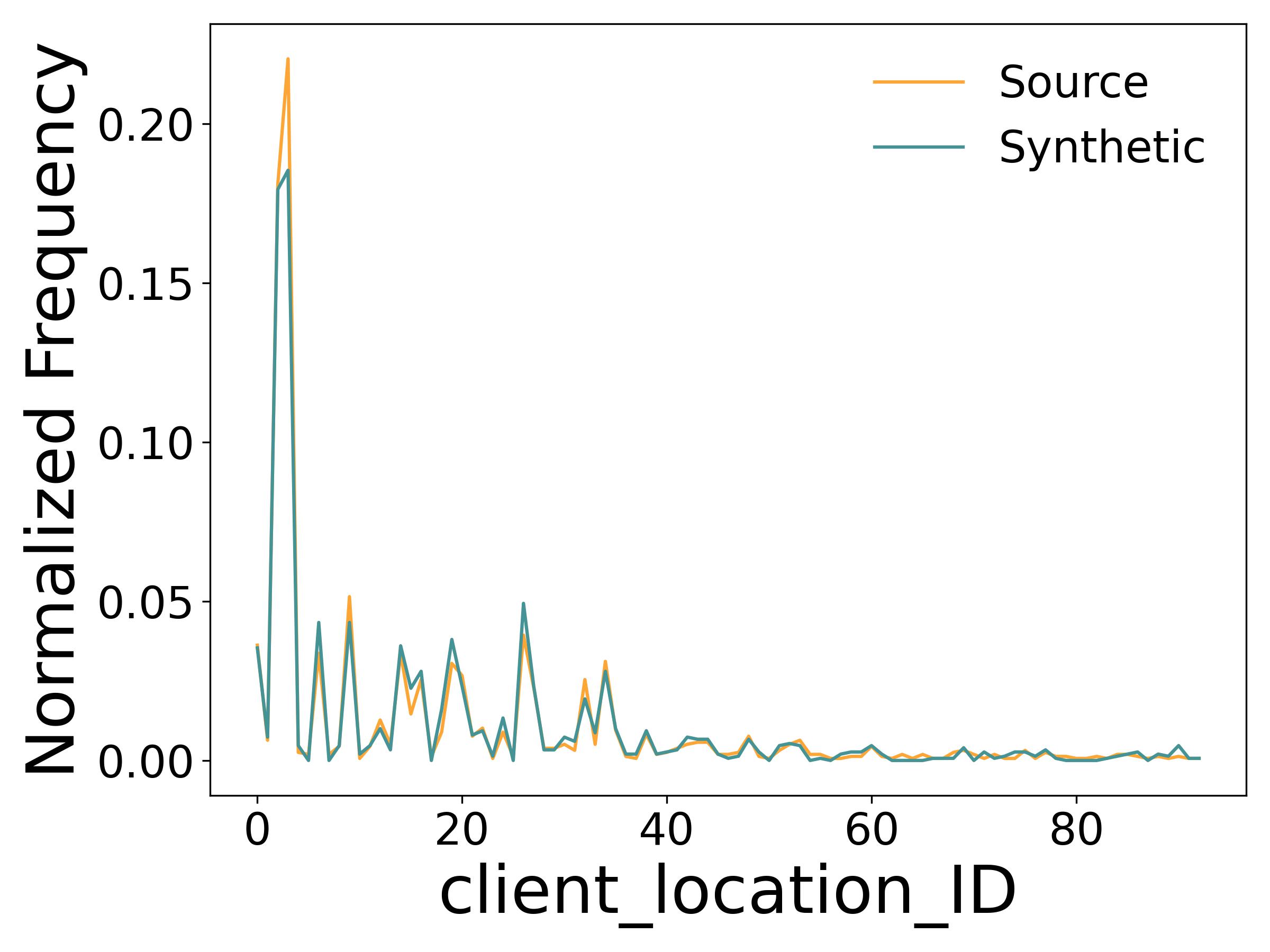}
    \subcaption{client\_location ID}
  \end{minipage}%
  \hfill
  \begin{minipage}{0.2\textwidth}
    \centering
    \includegraphics[width=\linewidth]{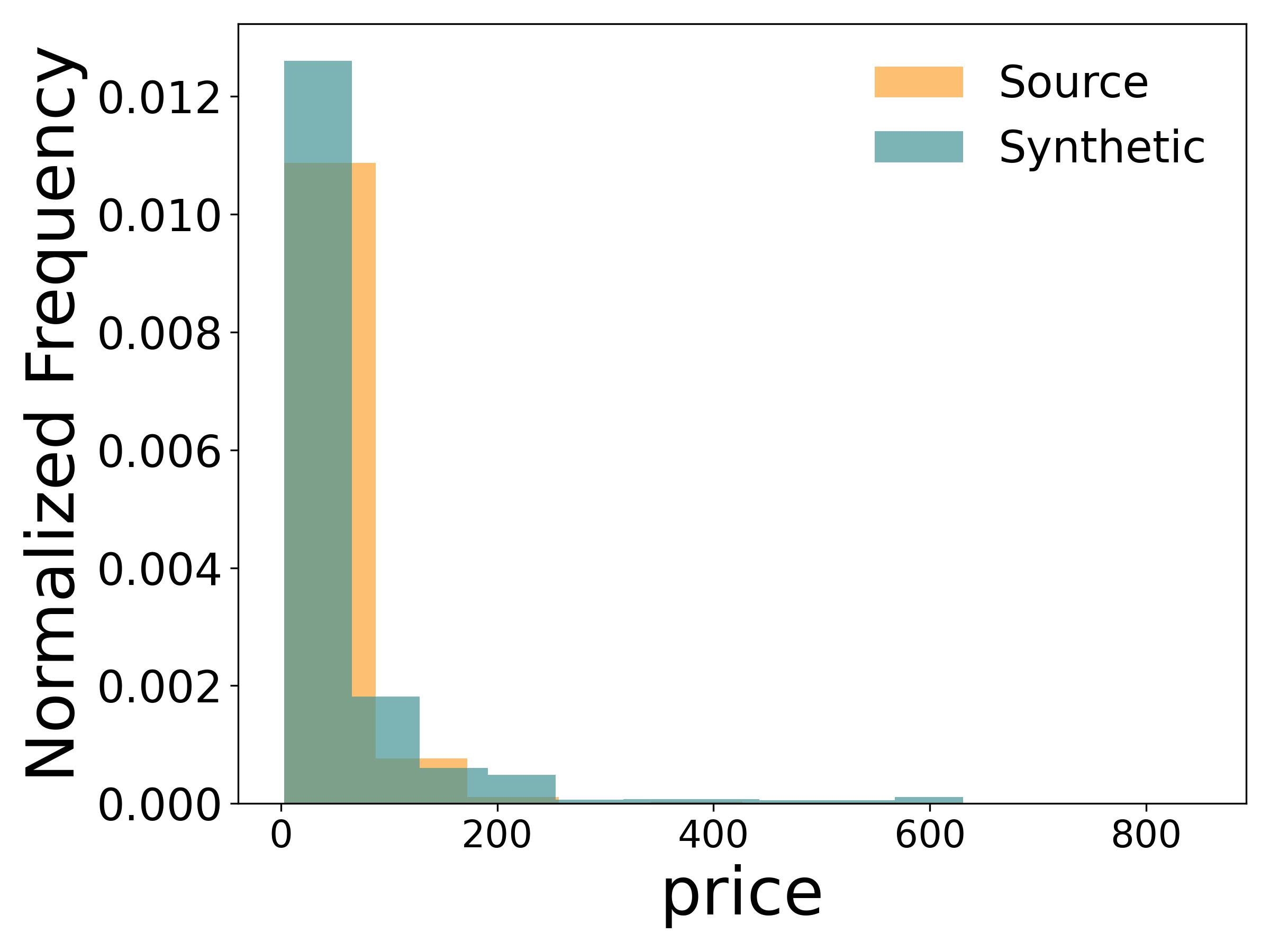}
    \subcaption{fixed\_price}
  \end{minipage}%
  \hfill
  \begin{minipage}{0.2\textwidth}
    \centering
 \includegraphics[width=\linewidth]{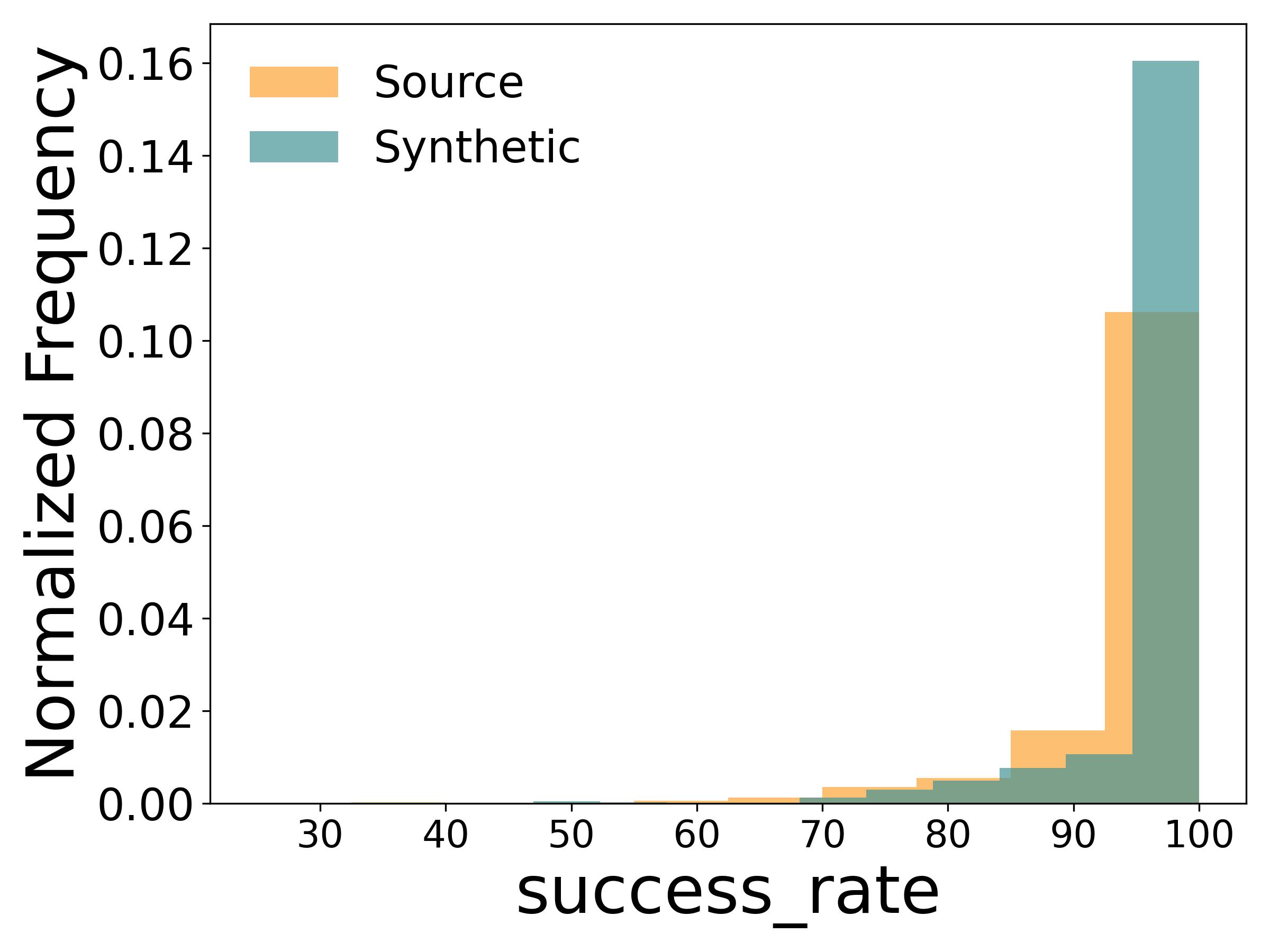}
    \subcaption{success\_rate}
  \end{minipage}
  \caption{Quality of the data related to the remaining attributes of \textit{worker-data}}
  \label{worker_qua}
\end{figure}

\begin{figure}[!t]
\vspace{-0.1in}
  \centering
  \includegraphics[width=0.8\columnwidth]{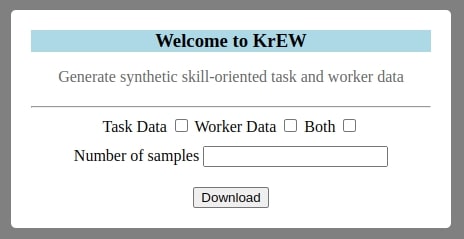}
  \caption{CTG-KrEW application's user interface}
  \label{tool_pic}
  \vspace{-0.2in}
\end{figure}

\section{Discussion and Use-Cases}
\label{dis}
\vspace{-0.1in}
The CTG-KrEW framework introduces a novel approach to generating synthetic tabular data by addressing critical challenges in maintaining the semantic integrity of contextually associated word sequences. The integration of word2vec with K-Means clustering allows CTG-KrEW to preserve meaningful relationships within the data, ensuring that the generated synthetic datasets are both realistic and contextually accurate. While initially designed for skill-oriented datasets, CTG-KrEW's capabilities extend far beyond this specific application, making it a versatile tool for a wide range of data generation tasks. Some of the practical uses cases of CTG-KrEW's application are:
\begin{itemize}
    \item \textbf{Customer Review Synthesis:} CTG-KrEW can generate synthetic datasets of customer reviews, preserving the contextually linked sequences of product features, customer sentiments, and usage scenarios. This synthetic data can be used for training sentiment analysis models or enhancing recommendation systems without compromising real customer data.
    \item  \textbf{Medical Report Generation:} In healthcare, CTG-KrEW can be used to generate synthetic medical reports, maintaining the associations between symptoms, diagnoses, and treatment plans, along with patient demographics and lab results. This enables the development and testing of predictive models for disease diagnosis and treatment planning while ensuring patient privacy.
    \item \textbf{Legal Document Synthesis:} CTG-KrEW can generate synthetic legal documents, such as contracts or case briefs, preserving the contextual relationships between legal terms, clauses, and conditions. This data can be used to train NLP models for legal text analysis or contract clause identification, aiding in the development of legal tech tools without risking confidentiality.
    \item \textbf{Financial Data Generation:} In the finance sector, CTG-KrEW can be employed to generate synthetic financial datasets that maintain the relationships between transaction descriptions, account types, and transaction amounts. This data is invaluable for developing models for fraud detection, risk assessment, or financial forecasting, all while keeping real financial data secure.
\end{itemize}

\vspace{-0.1in}
\section{Conclusion}
\label{conc}
\vspace{-0.1in}
The limitations of the generic CTGAN model, particularly in handling contextually associated word sequences and its high computational demands, prompted the development of the CTG-KrEW framework. By incorporating three key preprocessing steps, unique skill identification, word2vec encoding, and K-Means clustering—CTG-KrEW enhances the core CTGAN's ability to generate realistic synthetic tabular data. Experimental evaluations demonstrate that CTG-KrEW not only outperforms baseline methods but also significantly reduces memory usage and CPU time, making it a highly efficient solution.

Moreover, the development of the KrEW web-application underscores the practical utility of this framework, providing users with a freely accessible platform to generate synthetic data at any scale, tailored to the specific features of the training datasets. While this study primarily focused on skill-oriented datasets, CTG-KrEW's versatility extends far beyond this niche, offering robust capabilities for generating a wide range of categorical, numeric, and contextually associated data types.

Looking ahead, the framework's potential for broader application is immense. Future research will focus on expanding CTG-KrEW to support more complex and contextually nuanced data, such as phrases with specific tones, and exploring its application across various domains, thereby further solidifying its role as a comprehensive solution for synthetic data generation.
\vspace{-0.1in}
\section*{Competing Interest}
\vspace{-0.1in}
The authors declare that they have no known competing financial interests or personal relationships that could have appeared to influence the work reported in this article.

\vspace{-0.1in}
\section*{Funding}
\vspace{-0.1in}
No funding has been received for this research work.

\vspace{-0.15in}
\section*{Data availability}
\vspace{-0.1in}
The datasets used or analyzed during the study are available from the corresponding author upon reasonable request.

% \vspace{-0.1in}
\bibliographystyle{elsarticle-num}
\bibliography{sample}

\end{document}